\documentclass[10pt,twocolumn,letterpaper]{article}

\usepackage{iccv}              

\usepackage[hyphens]{url}
\usepackage{graphicx}
\usepackage{multirow}
\usepackage{booktabs}
\usepackage{bbding}
\usepackage{float}                  

\usepackage{enumitem}
\usepackage{zhlipsum}
\usepackage{colortbl}
\usepackage{enumitem}
\usepackage{mathtools,amsmath,amsfonts,bm}
\usepackage{fix-cm}

\usepackage[accsupp]{axessibility}

\def\ie{\emph{i.e.,~}}
\def\eg{\emph{e.g.,~}}

\newcommand{\supp}{\emph{supplementary material}}

\ifdefined \GramaCheck
  \newcommand{\CheckRmv}[1]{}
  \newcommand{\figref}[1]{Figure 1}%
  \newcommand{\tabref}[1]{Table 1}%
  \newcommand{\secref}[1]{Section 1}
  \newcommand{\algref}[1]{Algorithm 1}
  \renewcommand{\eqref}[1]{Equation 1}
\else
  \newcommand{\CheckRmv}[1]{#1}
  \newcommand{\figref}[1]{Figure~\ref{#1}}
  \newcommand{\tabref}[1]{Table~\ref{#1}}
  \newcommand{\secref}[1]{\S\ref{#1}}
  \renewcommand{\eqref}[1]{(\ref{#1})}
\fi

\definecolor{citecolor}{HTML}{0071BC}
\definecolor{linkcolor}{HTML}{ED1C24}

\newcommand{\sArt}{{state-of-the-art~}}
\newcommand{\ourframework}{DynamicVoyager\xspace}

\definecolor{jinqidarkred}{HTML}{990000}

\newcommand\calP{{\mathcal{P}}}

\newcommand\mD{{\boldsymbol{D}}}

\newcommand\mI{{\boldsymbol{I}}}

\newcommand\mM{{\boldsymbol{M}}}

\newcommand\mhD{{\boldsymbol{\hat{D}}}}

\newcommand\mhI{{\boldsymbol{\hat{I}}}}

\newcommand\mhM{{\boldsymbol{\hat{M}}}}

\newcommand\vc{{\boldsymbol{c}}}

\newcommand\vo{{\boldsymbol{o}}}
\newcommand\vp{{\boldsymbol{p}}}

\newcommand\vr{{\boldsymbol{r}}}

\newcommand\vx{{\boldsymbol{x}}}

\newcommand\bPi{\boldsymbol{\Pi}}

\newcommand\sR{{\mathbb{R}}}

\newcommand{\myparagraph}[1]{\smallskip\noindent\textbf{#1}}

\newcommand\mbD{{\boldsymbol{\bar{D}}}}

\definecolor{iccvblue}{rgb}{0.21,0.49,0.74}
\usepackage[pagebackref,breaklinks,colorlinks,allcolors=iccvblue]{hyperref}
\usepackage{inconsolata}

\def\paperID{3989} 
\def\confName{ICCV}
\def\confYear{2025}

\title{Voyaging into Perpetual Dynamic Scenes from a Single View}
\author{Fengrui Tian \quad Tianjiao Ding \quad Jinqi Luo \quad Hancheng Min \quad Ren\'e Vidal\\
University of Pennsylvania\\
{\tt\small \{tianfr,tjding,jinqiluo,hanchmin,vidalr\}@upenn.edu}
}

\begin{document}
\twocolumn[{%
\maketitle
    \begin{center}
		\centering
        \CheckRmv{
        \captionsetup{hypcap=false}
    \includegraphics[width=.98\linewidth]{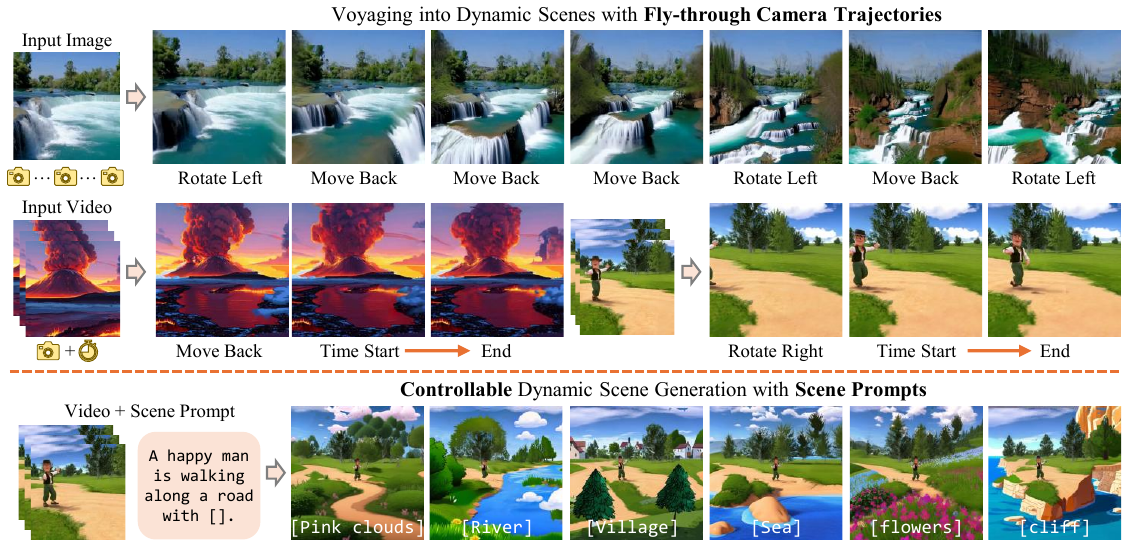}
\captionof{figure}{
\ourframework~generates 4D point clouds of unbounded dynamic scenes by our dynamic scene outpainting process. Given a fixed viewpoint video (or an image with a motion prompt) with dynamic scene prompts and fly-through camera trajectories, \ourframework can generate dynamic scenes along the trajectories (\textbf{Top}) and control scene generation contents (\textbf{Bottom}).
	}
}
\label{fig:teaser}
\end{center}	
\label{fig:teaser_main}
}]

\begin{abstract}
The problem of generating a perpetual dynamic scene from a single view is an important problem with widespread applications in augmented and virtual reality, and robotics. However, since dynamic scenes regularly change over time, a key challenge is to ensure that different generated views be consistent with the underlying 3D motions. Prior work learns such consistency by training on multiple views, but the generated scene regions often interpolate between training views and fail to generate perpetual views. To address this issue, we propose \ourframework, which reformulates dynamic scene generation as a scene outpainting problem with new dynamic content. As 2D outpainting models struggle at generating 3D consistent motions from a single 2D view, we enrich 2D pixels with information from their 3D rays that facilitates learning of 3D motion consistency. More specifically, we first map the single-view video input to a dynamic point cloud using the estimated video depths. We then render a partial video of the point cloud from a novel view and outpaint the missing regions using ray information (e.g., the distance from a ray to the point cloud) to generate 3D consistent motions. Next, we use the outpainted video to update the point cloud, which is used for outpainting the scene from future novel views. Moreover, we can control the generated content with the input text prompt. Experiments show that our model can generate perpetual scenes with consistent motions along fly-through cameras. Project page:
\url{https://tianfr.github.io/DynamicVoyager}.
\end{abstract}    
\section{Introduction}
\label{sec:intro}
Perpetual scene generation \cite{21iccv/liu_infinite,22eccv/li_infinitzero,24cvpr/yu_wonderjourney} aims to create a virtual 3D scene as the camera moves along arbitrary trajectories, typically starting from a single view observation.
While recent studies \cite{22eccv/li_infinitzero,24cvpr/yu_wonderjourney,24arxiv/yu_wonderworld} have achieved significant progress in perpetual generation of \textit{static} scenes by leveraging image outpainting models \cite{22cvpr/rombach_ldm,25arxiv/flux,flux2024}, these approaches cannot generate scenes with \textit{dynamic} content (\eg waving hands, flowing rivers). Such dynamic content has important applications in augmented reality (AR), virtual reality (VR) \cite{15tog/alvaro_freevideo,20tog/michael_immersive} and robotics \cite{21nips/gan_tdw}. For example, AR/VR game designers need to build a perpetual dynamic scene for players to explore and interact, while roboticists use virtual scenes containing commonly seen dynamic objects in natural environments for training embodied agents \cite{24arxiv/yang_thinkspc,23iclr/gu_maniskill} by self-exploration. 

A key challenge in generating perpetual dynamic scenes is to ensure that the generated dynamic content has 3D consistent motions: \ie the motions observed in any two views must correspond to the same underlying 3D dynamics.
Generating consistent motions from a single view is inherently ambiguous due to the limited 3D motion information contained in a 2D image. Previous dynamic scene generation methods \cite{24arxiv/xie_sv4d,24arxiv/sun_dimensionx,24nips/yu_4real,25iclr/4k4dgen,24cvpr/ling_ayg} address this challenge by learning from multiple views surrounding the scenes so that 3D motion information can be implicitly learned from the 2D motions observed in these views. However, such a learning strategy severely restrains models from creating perpetual dynamic content, because the generated dynamic regions often interpolate between training views and fail to generate perpetual views, as exemplified in \figref{fig:prev_failure}.

In this paper, we propose \ourframework, a novel approach to perpetual dynamic scene generation that reformulates the task as a scene outpainting problem, enabling the synthesis of new dynamic content in previously unseen regions of the scene. Rather than relying on multiple 2D views as input,  \ourframework learns 3D-consistent motion dynamics from single-view observations by treating \textit{pixels as rays}, thereby enriching each pixel with the contextual information of its corresponding camera ray.

More specifically, given an input image, we exploit image-to-video diffusion models \cite{24iclr/yang_cogvideox,22iclr/hong_cogvideo} to synthesize a fixed pose video from that image. To initialize a dynamic scene from the video input, we estimate the video depth maps and backproject the fixed viewpoint video frames into 3D space as dynamic point clouds with the estimated depths. After that, we move the camera to a partially unseen area of the scene and rasterize the reconstructed point clouds into an incomplete video from that view. To outpaint the video with 3D consistent motions, we consider pixels as rays to complement the video input with 3D scene information. For pixels in the visible area, we obtain 3D motion information by sampling ray depth maps from the point clouds. For pixels in the unseen area, we backproject rays from pixels and use the distance between these rays and the point cloud to infer the 3D spatial relationship between the unseen area and the visible area. We then outpaint the incomplete video with the sampled ray depth and ray distance maps for the visible and unseen parts, respectively. Finally, we estimate the depth maps of the outpainted video and update the unseen area of the dynamic point clouds. \figref{fig:teaser} visualizes examples of dynamic scenes generated by our method using input from a video with a fixed camera pose or a single image. It also shows that we can control the perpetual generation with input scene prompts.

\CheckRmv{
\begin{figure}[t]
  \centering
   \includegraphics[width=\linewidth]{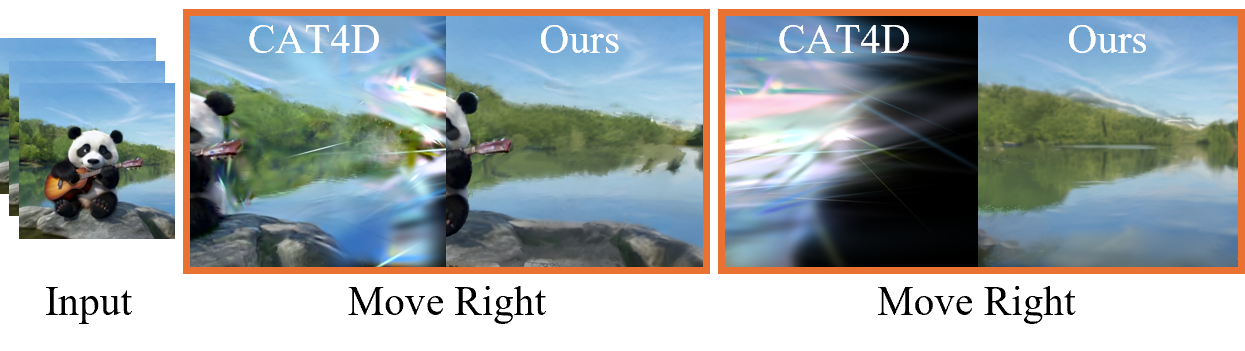}

   \caption{Failure examples of previous dynamic scene generation methods. While the generated dynamic scenes from previous works \cite{24arxiv/sun_dimensionx,24arxiv/cat4d,24arxiv/xie_sv4d} are strictly bounded by the input views, \ourframework successfully generates dynamic scenes with large camera motions by the proposed scene outpainting process.}
   \label{fig:prev_failure}
\end{figure}
}
To summarize, our contributions are as follows:
\begin{enumerate}
    \item We reformulate the dynamic scene generation problem as a scene outpainting problem, so that our generated scene can be explored from a single view to any place through fly-through camera trajectories.
    \item We generate view-consistent motions in 3D space by treating pixels as rays, which allows us to use 3D information to enrich the outpainting model.
    \item We present experiments showing that, unlike other bounded scene generation methods, our model can generate perpetual dynamic scenes and control the generation with scene prompts.
\end{enumerate}
\section{Related Work}
\label{sec:related_work}
\CheckRmv{
\begin{figure*}[t]
  \centering
   \includegraphics[width=\linewidth]{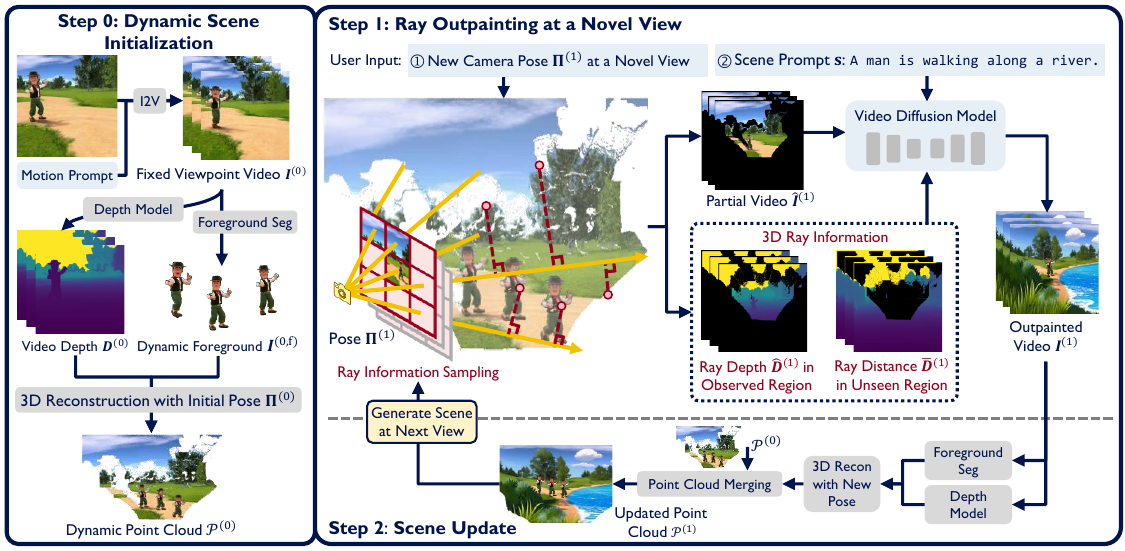}

   \caption{The overview of \ourframework. First, for an initial camera pose, we build the dynamic point clouds from the input image by employing the image-to-video diffusion model \cite{24iclr/yang_cogvideox}, depth model \cite{20pami/rene_midas} and foreground segmentation model \cite{23cvpr/jitesh_oneformer} (\secref{sec:approach_step0}). Then given a new camera pose, we render the partial video and the corresponding ray depth and ray distance maps. We consider pixels as rays to enrich the pixel information of the partial video with the corresponding ray depth and ray distance information for outpainting the video with consistent motions (\secref{sec:approach_step1}). Finally, we update the dynamic point cloud with the outpainted video and the corresponding video depth (\secref{sec:approach_step2}).}
   \label{fig:overview}
\end{figure*}
}
\myparagraph{Static scene generation.}
With the development of vision foundation models \cite{23iccv/alexander_sam,24cvpr/wang_dust3r,20pami/rene_midas,23cvpr/jitesh_oneformer,22cvpr/robin_sd}, many works have started to generate static scenes from input images \cite{24nips/gao_cat3d,22nips/miguel_gaudi,21iccv/terrance_unconstrained} or text prompts \cite{24iclr/liu_syncdreamer,23iccv/lukas_t2room}. Early efforts
focused on indoor scene generation \cite{23iccv/lukas_t2room,21iccv/terrance_unconstrained,22nips/miguel_gaudi,23cvpr/lei_rgbd2} or object-centric scene generation \cite{24nips/gao_cat3d}. These methods firstly generate multiview images of the target scene and fuse the multiview images by using 3DGS \cite{23tog/bernhard_3dgs} or NeRF \cite{20eccv/Ben_nerf,23iccv/tian_mononerf,24iclr/tian_semflow} representations. However, they can only generate static scenes with limited ranges of camera motions. When the camera moves drastically, these methods fail to generate unseen regions with consistent and clear scene structures. To overcome this issue, later studies have started to explore \textit{perpetual scene generation} \cite{21iccv/liu_infinite} that allows larger camera movements \cite{253dv/popov_camctrl3d,24cvpr/yu_wonderjourney,24arxiv/yu_wonderworld,21iccv/liu_infinite,22eccv/li_infinitzero,21iccv/hu_worldsheet}. Notably, WonderJourney \cite{24cvpr/yu_wonderjourney} and WonderWorld \cite{24arxiv/yu_wonderworld} propose to generate an infinite world with diverse contents by using a combination of \sArt depth estimation \cite{20pami/rene_midas}, image segmentation \cite{23iccv/alexander_sam} and object detection models \cite{23cvpr/jitesh_oneformer}. However, these methods can only generate diverse scenes with static objects. In this paper, we target the problem of generating dynamic scenes with new dynamic contents with fly-through camera trajectories.

\myparagraph{Dynamic scene generation.}
Following the great success of static scene generation methods, recent studies have started to address the dynamic scene generation problem \cite{24arxiv/pan_efficient4d,23arxiv/yin_4dgen,23arxiv/ren_dreamgs4d,24cvpr/ling_ayg,24arxiv/xie_sv4d,24arxiv/sun_dimensionx,25iclr/4k4dgen,24arxiv/cat4d,24nips/yu_4real}. Due to the limited resources of real-world dynamic scene data, many researchers try to leverage the pretrained video diffusion models \cite{24iclr/yang_cogvideox,22iclr/hong_cogvideo,24eccv/xing_dyncrafter,25cvpr/ren_gen3c} to generate multiple views of dynamic scenes. A key challenge in this setting is maintaining 3D motion consistency in multiple generated views.  CAT4D \cite{24arxiv/cat4d} and DimensionX \cite{24arxiv/sun_dimensionx} learn to generate 3D consistent motion from multiview datasets \cite{24cvpr/lu_dl3dv,23cvpr/matt_objaverse}. However, as the multiview images in these datasets are taken from cameras that are very close to each other, the generated dynamic scenes in these methods are strictly limited by the input images or videos. 4K4DGen \cite{25iclr/4k4dgen} animates 3D consistent motion by building a 3D consistent noise space in the diffusion models. GEN3C \cite{25cvpr/ren_gen3c} achieves camera control of dynamic scenes with 4D point cloud representations. However, since these methods only interpolate novel views surrounding the input image or videos, they still cannot generate new dynamic contents in the unseen regions of the input images.  In this paper, we consider dynamic scene generation as a scene outpainting problem so that our method can generate new dynamic content with fly-through camera motions in the unseen regions of the input video.

\myparagraph{Video diffusion models.}
Our method relies on the power of the recent video diffusion models \cite{24eccv/xing_dyncrafter,24iclr/yang_cogvideox} to outpaint dynamic scenes with diverse contents. While some researchers propose video diffusion models with controllable camera trajectories \cite{25iclr/sherwin_vd3d,24iclr/guo_animatediff,25iclr/he_camctrl,25iclr/hou_camctrlfree,24nips/kuang_collcamctrl,24eccv/van_gencamdolly,24siggraph/yang_directavideo,25iclr/zhao_genxd}, these trajectories are bounded by the regions of the input views (a failure case is shown in \figref{fig:motionctrl_comp}). Besides, since the outputs of these methods are generated videos without any 4D scene structures, they always encounter view inconsistency issues when fusing 4D representation with the output videos. Instead, our method adopts 4D point clouds as the dynamic scene representation and outpaints the dynamic point clouds with video outpainting models, where the unseen regions in the input video can be generated by the outpainting procedure, and the motion consistency in different views can be achieved by introducing ray information in 3D space.

\section{Approach}
\label{sec:approach}

In this section, we present the proposed \ourframework approach for generating perpetual dynamic scenes, which is summarized in \figref{fig:overview}. \ourframework consists of three steps, described in \secref{sec:approach_step0}, \secref{sec:approach_step1} and \secref{sec:approach_step2}, respectively: 
\begin{itemize}[left=0pt,itemsep=0pt,parsep=0pt,topsep=5pt]
    \item \emph{Dynamic Scene Initialization:} \ourframework accepts either a video with a fixed camera pose or an image with a prompt describing object motions as input---in the latter case, it first generates a video with a fixed pose.
    A depth map is then estimated from each frame of the video. 
    We segment the video into a dynamic foreground and a static background, reconstruct the point clouds separately, and aggregate these into one dynamic point cloud.
    \item \emph{Ray Outpainting:} At a new given pose, the goal is to generate a video consistent with the existing dynamic point cloud and a given scene prompt. We first rasterize the point cloud at the new pose, which is consistent by construction. That said, the rasterization only fills in part of the image, so the question is how to outpaint the rest \textit{consistently}. To do so, we propose to guide the outpainting process with 3D information by providing the model with ray depth in the observed region and the distance of rays to the point cloud in the unseen region. 
    \item \emph{Dynamic Scene Update:} We update the dynamic point cloud using the depth maps of the video frames from the new pose. We repeat the steps in \secref{sec:approach_step1} and \secref{sec:approach_step2} to iteratively update the dynamic point cloud from novel poses.
\end{itemize}


\subsection{Dynamic Scene Initialization}
\label{sec:approach_step0}
We begin by describing how to generate an initial dynamic point cloud given a video at a fixed camera pose $\bPi^{(0)}$ or an image with a motion prompt. Notation-wise, a superscript $(i)$ denotes objects related to the $i^\mathrm{th}$ camera pose $\bPi^{(i)}$.  

For illustration,
suppose we are given an image $\mI_0^{(0)} \in \sR^{h\times w \times 3}$, where $h, w, 3$ are respectively height, width and number of color channels, as well as a prompt describing the desired motion for the video.  To generate a video at the fixed pose $\bPi^{(0)}$, we employ a pretrained image-to-video diffusion model \cite{24iclr/yang_cogvideox} by providing both $\mI_0^{(0)}$ and the motion prompt, where the motion prompt is further prepended with \texttt{Camera is strictly fixed}. We let  $\mI^{(0)} = \{\mI^{(0)}_t\}_{t=0}^{N-1}$ denote the generated video, where $\mI^{(0)}_t$ denotes its $t^{\mathrm{th}}$ frame. 



Given the generated or user-input fixed viewpoint video $\mI^{(0)}$, we reconstruct the underlying dynamic scene, represented as a 4D point cloud $\mathcal{P}^{(0)}=\{ \vp =(\vx,t,\vc)\}$, where $\vx\in \mathbb{R}^3$, $t\in \mathbb{R}$ and $\vc\in \mathbb{R}^3$ denote the 3D position, timestamp and color respectively. Since the objects in the video foreground and the scene in the background have different dynamic natures,  our framework reconstructs point cloud $\mathcal{P}^\mathrm{(0,f)}$ for the foreground scene and $\mathcal{P}^\mathrm{(0,b)}$ for background separately. 
To this end, we extract the binary video foreground masks $\mM^{(0)}=\{\mM^{(0)}_t\in \{0,1\}^{h\times w}\}_{t=0}^{N-1}$ with the foreground segmentation method \cite{23cvpr/jitesh_oneformer} to segment the video $\mI^{(0)}$ into foreground video $\mI^{(0,\mathrm{f})} = \{\mI^{(0,\mathrm{f})}_t\}_{t=0}^{N-1}$  and background video $\mI^{(0,\mathrm{b})} = \{\mI^{(0,\mathrm{b})}_t\}_{t=0}^{N-1}$ .
We employ the depth model \cite{20pami/rene_midas} on video $\mI^\mathrm{(0)}$ to obtain the video depth maps $\mD^{(0)} = \{\mD^{(0)}_t\in \mathbb{R}^{h\times w}\}_{t=0}^{N-1}$. 
Next, we describe the foreground and background scene initialization steps.

\myparagraph{Scene foreground initialization.} 
We exploit the depth maps $\mD^{(0)}$ and foreground masks $\mM^{(0)}$ to obtain the foreground depth maps $\mD^{(0,\mathrm{f})} =  \{\mD^{(0,\mathrm{f})}_t\}_{t=0}^{N-1}$. Since the video is captured with a fixed camera, we initialize the video camera pose $\bPi^{(0)}$ as the center of the dynamic scene. We reconstruct the initial foreground point cloud by employing the foreground video frames $\mI^{(0,\mathrm{f})}$, foreground depth maps $\mD^{(0,\mathrm{f})}$ and the initial camera pose $\bPi^{(0)}$,
\begin{equation}
\label{eq:mapping}
    \calP_t^{(0,\mathrm{f})} = \phi ([\mI_t^{(0,\mathrm{f})}, \mD_t^{(0,\mathrm{f})}], \bPi^{(0)}, t), \ t\! \in\! \{0,...,N\!-\!1\}\,,
\end{equation}
where $\calP_t^{(0,\mathrm{f})}$ denotes the foreground point cloud at timestamp $t$.  $\mathbf{\phi}$ denotes the mapping from an image with its depth map to the point cloud. We obtain the foreground point cloud at all timestamps $\calP^{(0,\mathrm{f})}=\bigcup_{t=0}^{N-1}\calP_t^{(0,\mathrm{f})}$. 

\CheckRmv{
\begin{figure}[t]
  \centering
   \includegraphics[width=\linewidth]{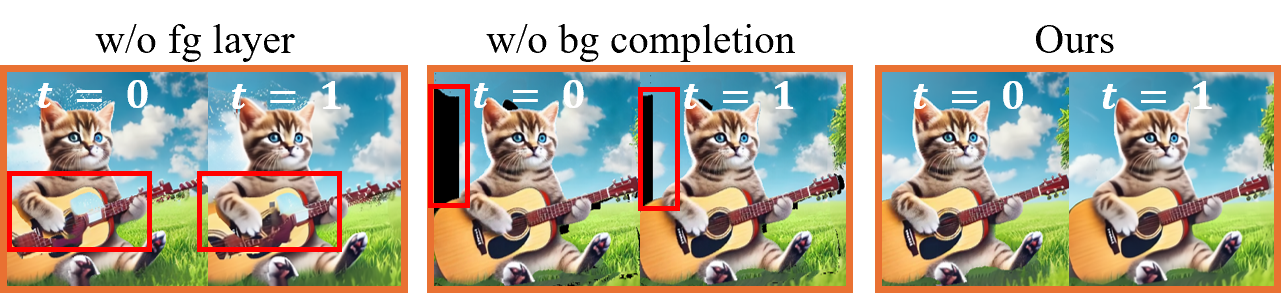}
   \caption{Qualitative ablation studies of the proposed foreground layer (\S \ref{sec:approach_step0}) and background completion (\S \ref{sec:approach_step2}) strategy.}
   \label{fig:ablation_4d_representation}
\end{figure}
}

\myparagraph{Scene background initialization.}
Although the background geometry of a dynamic scene remains static over time, its appearance can vary across frames due to dynamic textures such as flowing water, moving clouds, or changing lighting. Hence, we model the background as having constant depth but varying color across time. Also, notice that foreground objects may occlude parts of the background at certain timestamps, making some background pixels unobservable at those frames. To address this, we leverage foreground masks to identify non-occluded regions and compute the background depth at each pixel location. Specifically, let $\mD^{(0)}_t(x,y)$ and $\mM^{(0)}_t(x,y)$ denote the depth and foreground mask values at pixel $(x,y)$ of the video frame $\mI^{(0)}_t$. The background depth can be computed element-wise as
\begin{equation}
\label{eq:smooth}
    \mD^{(0,\mathrm{b})}(x,y) \!=\! \frac{\sum_{t=0}^{N\!-\!1}\mD^{(0)}_t(x,y)\!\cdot\!\big(1\!-\! \mM^{(0)}_t(x,y)\big)}{\sum_{t=0}^{N\!-\!1}\big(1\!-\! \mM^{(0)}_t(x,y)\big)}\,,
\end{equation}
where $\mD^{(0,\mathrm{b})}$ is the refined background depth map.
We employ the same mapping function $\phi$ to map the background regions on the video frames to the 4D point clouds,
\begin{equation}\calP_t^{(0,\mathrm{b})\!}=\!\phi([\mI_t^{(0,\mathrm{b})}, \mD^{(0,\mathrm{b})}],\bPi^{(0)}, t), t\! \in\! \{0,...,N\!-\!1\}.
\end{equation}
In this way, the obtained background point clouds $\calP^{(0,\mathrm{b})}=\bigcup_{t=0}^{N-1}\calP_t^{(0,\mathrm{b})}$ have time-invariant positions and time-varying colors.
Finally, we obtain the scene point cloud by merging the foreground and background point clouds,
\begin{equation}
    \calP^{(0)} = \calP^{(0,\mathrm{f})}\cup \calP^{(0,\mathrm{b})},
\end{equation}
which is then introduced for outpainting the scene at a novel view with 3D consistent motions.
\subsection{Ray Outpainting at a Novel View}
\label{sec:approach_step1}
To outpaint the scene at a given novel pose $\bPi^{(1)}$,
we follow the assumption of previous works \cite{24arxiv/yu_wonderworld,24cvpr/yu_wonderjourney,24eccv/kong_dreamdrone} that the view of $\bPi^{(1)}$ overlaps moderately with that of $\bPi^{(0)}$ to allow for outpainting a substantial portion of the scene. 
The question is, how can we outpaint a dynamic scene containing consistent motions with the existing dynamic point cloud $\mathcal{P}^{(0)}$? 

To address this question, we first rasterize the point cloud at $\bPi^{(1)}$. 
As shown in the top row of \Cref{fig:ablation_ray}, we obtain a partial video whose observed region is the projection of the portion of the point cloud visible from $\bPi^{(1)}$.
To handle the unseen region, a naive approach would be to directly outpaint the partial video. However, this typically leads to inconsistencies across the boundary of observed and unseen regions, as shown in \Cref{fig:ablation_ray}. Our intuition is that pixels in each video frame only provide 2D information, which is not enough to reconstruct the 3D motion. 
To address this issue, we consider the 3D ray from the camera origin through each pixel. If the pixel is in the observed region, we compute its depth. If the pixel is in the unseen region, we compute the distance from its ray to the point cloud.
Both the depth and distance maps serve as guidance for how to outpaint a pixel while respecting information from the existing point cloud.

\myparagraph{Ray information sampling in observed region.}
Given a novel pose $\bPi^{(1)}$, we first rasterize the dynamic point cloud $\mathcal{P}^{(0)}$:
at every time step $t\in\{0,\cdots,N-1\}$, we compute
\begin{equation}
    (\mhI^{(1)}_t, \mhD^{(1)}_t,\mhM^{(1)}_t)= \varphi(\mathcal{P}^{(0)}_t, \bPi^{(1)}).
\end{equation}
Here, $\varphi$ denotes the image rasterization function. $\mhM^{(1)}_t$ is a binary mask, where $\mhM^{(1)}_t(x,y)=1$ if at least one 3D point from $\mathcal{P}^{(0)}_t$ hits the image at pixel $(x,y)$
during rasterization and $0$ otherwise. Correspondingly, $\mhI^{(1)}_t$ is the rasterized partial video and $\mhD^{(1)}_t$ the rasterized \textit{ray depth} map (to be distinguished from the depth map generated by depth models as in \S \ref{sec:approach_step0}). A hat on the notation emphasizes that the signals are not observed for all locations, and we denote $\mhD^{(1)}=\{\mhD^{(1)}_t\}_{t=0}^{N-1}$, $\mhM^{(1)}=\{\mhM^{(1)}_t\}_{t=0}^{N-1}$ and $\mhI^{(1)}=\{\mhI^{(1)}_t\}_{t=0}^{N-1}$. A few remarks are in order. First, by doing the rasterization, the observed regions are obviously consistent with the point cloud. Second, the ray depth maps encode the 3D information of the observed scene, which as we shall see will be included as guidance for outpainting. Third, for the regions not observed at $\bPi^{(1)}$, we introduce the distance between rays and the point cloud to complement the 3D information in these regions, as described next.

\CheckRmv{
\begin{figure}[t]
  \centering
   \includegraphics[width=\linewidth]{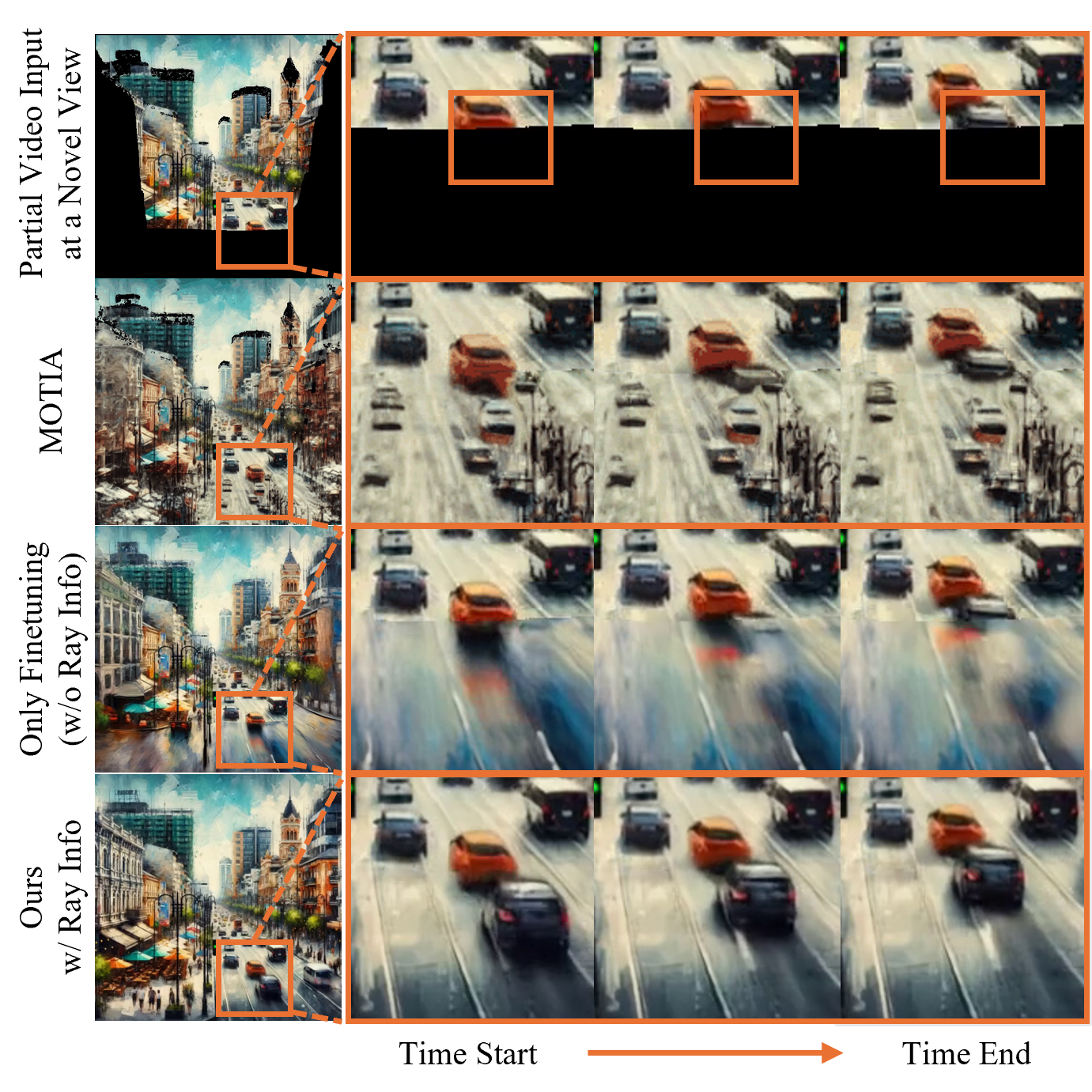}
   \caption{Ablation studies of scene outpainting with ray information. We also compare with the 2D outpainting model MOTIA \cite{24eccv/wang_motia}. Detailed visualization demonstrates that with ray information, our model successfully outpaints the dynamic scenes with consistent motions.}
   \label{fig:ablation_ray}
\end{figure}
}

\CheckRmv{
\begin{figure*}[t]
  \centering
   \includegraphics[width=\linewidth]{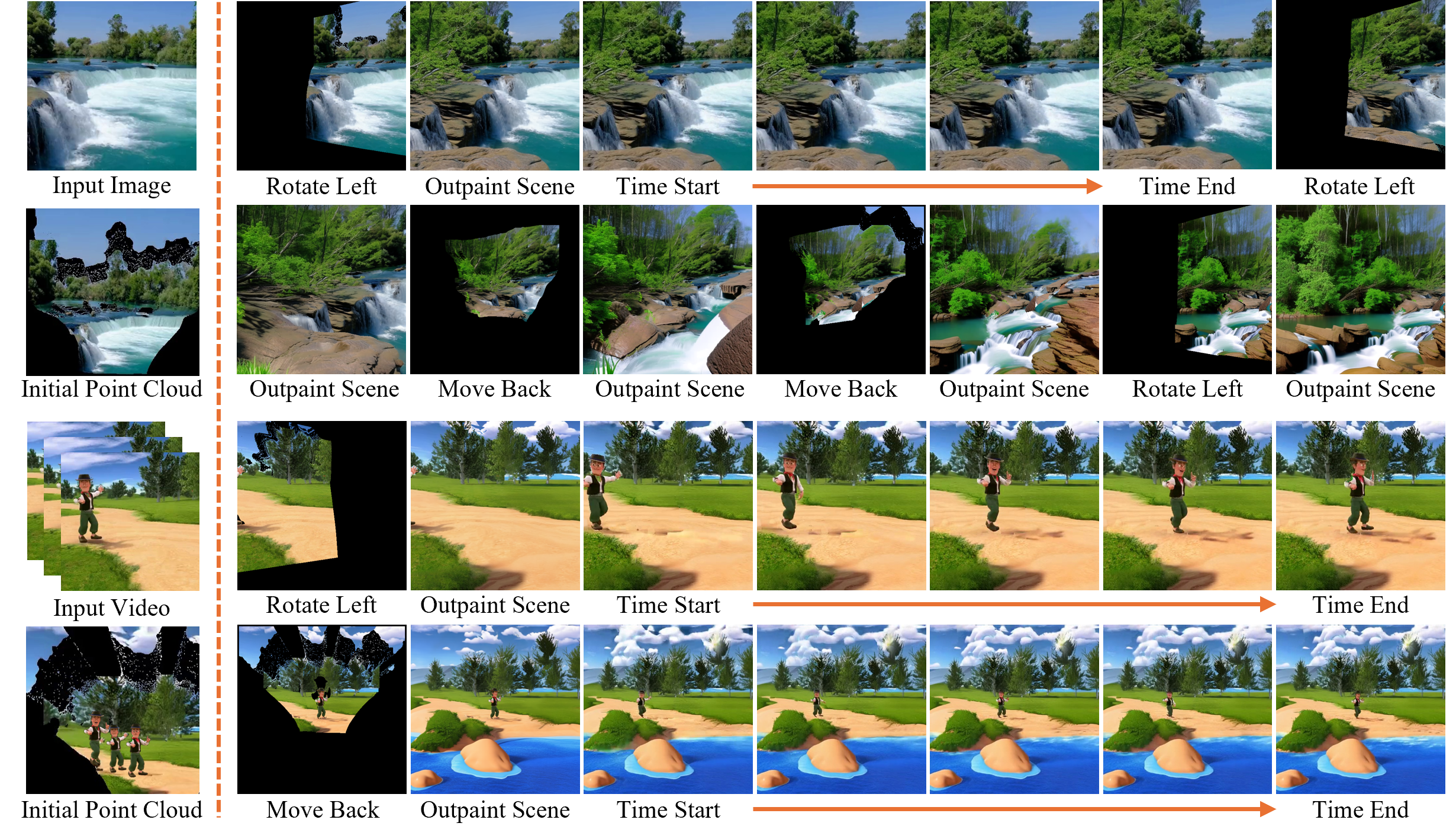}
   \caption{Voyaging into perpetual dynamic scenes with long camera trajectories from a single image or fixed viewpoint video. We iteratively show the unseen regions through large camera motions, the dynamic scene outpainting results and the consistent motions from novel views. By enriching the pixel information with ray contexts, our method successfully builds large dynamic scenes from input views.}
   \label{fig:large_cam_motion}
\end{figure*}
}
\myparagraph{Ray-point-cloud distance computation in unseen region.}
Fix the reference coordinate system at the origin of a camera. Let $\vr\in \sR^3$ be a unit vector denoting the direction of a ray from the origin pointing at a pixel location, and $\vp \in \sR^3$ a point in the 3D space. We recall that the distance between the ray and the 3D point can be computed as
\begin{equation}
    \operatorname{dist}_{\mathrm{r2p}}(\vr, \vp)=\sqrt{\|\vp\|_2^2 - (\vr^\top \vp)^2}.
\end{equation}
With the above said, we can compute the distance between camera rays and the 4D point cloud for the complementary 3D information of the unseen regions. Namely, we compute
\begin{equation}
\label{eq:r2p}
    \mbD^{(1)}_t(x,y) = \min_{\vp\in \mathcal{P}_t^{(0)}}\operatorname{dist}_{\mathrm{r2p}}\big(\vr^{(1)}(x,y), \vp-\vo^{(1)}\big),
\end{equation}
where $\vr^{(1)}(x,y) \in \mathbb{R}^3$ is the unit-norm ray vector starting from the origin of $\bPi^{(1)}$ pointing at pixel $(x,y)$, and $\vo^{(1)}$ is the camera center of $\bPi^{(1)}$,
both in the reference coordinate system of $\bPi^{(0)}$. As we will see, such a distance will be used to guide the outpainting.


\myparagraph{Ray outpainting.}
To outpaint the dynamic scene at the novel pose $\bPi^{(1)}$ from the partial video, we desire a few properties for the outpainted video: 1) the outpainted video needs to have a static pose fixed at $\bPi^{(1)}$ and be consistent with the existing dynamic point cloud in the observed region, 2) the user should be able to control the generation in the outpainted region (i.e., the unseen region), and 3) the motions in the outpainted region need to be consistent with the motions in the observed region.
To fulfill properties 1) and 2), we design our video outpainting model to take as input the extracted partial video $\mhI^{(1)}_t$, as well as a prompt narrating the desired motions in the generated scene. For property 3), we propose to employ ray depth map $\mhD^{(1)}$ and ray distance map $\mbD^{(1)}$ as guidance on 3D information in the observed region and how much the outpainted ray in the unseen region should respect the existing point cloud. To sum up, our video outpainting model accepts the partial video $\mhI^{(1)}$, scene prompt $\mathbf{s}$, ray depth map $\mhD^{(1)}$, ray distance map $\mbD^{(1)}$ and generates an outpainted video $\mI^{(1)} = \{\mI^{(1)}_t\}_{t=0}^{N-1}$. Next we describe how to train such a model.


\CheckRmv{
\begin{figure*}[!t]
  \centering
   \includegraphics[width=\linewidth]{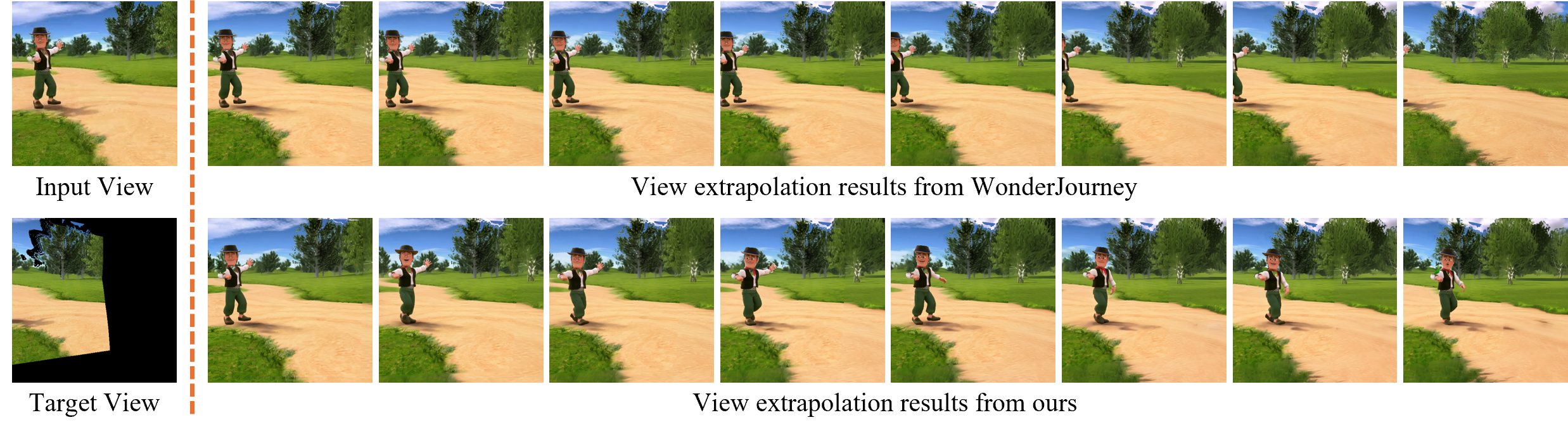}
   \caption{View extrapolation results. While Wonderjourney \cite{24cvpr/yu_wonderjourney} only extrapolates views of a static character in 3D space, \ourframework renders extrapolated views with 3D consistent motions by leveraging the ray information for scene outpainting.}
   \label{fig:view_interpolation}
\end{figure*}
}

\myparagraph{Training the outpainting model.} We collect a small video dataset for training (details in \supp), and reconstruct point clouds from it to compute ray information for supervision.
More specifically, 
given a training video $\mI$, we extract depth maps $\mD$ using the depth model \cite{20pami/rene_midas} and backproject the video from a camera $\bPi$ at the world center to reconstruct a point cloud.
We then move the camera closer to the scene and filter out the 3D points that are no longer visible from the current viewpoint. Then we move the camera back to the original camera pose $\bPi$ and compute a partial video $\mhI$, a partial depth map $\mhD$, and a ray distance map $\mbD$ following \secref{sec:approach_step1}.
To train the model, we add noise to $\mI$ to obtain $z_\tau$ at step $\tau$, and employ ControlNet \cite{23iccv/zhang_controlnet} to inject ray information.
Let $\theta_\mathrm{v}, \theta_\mathrm{c}$ denote the training parameters of the video diffusion model and ControlNet model. The training objective is
\begin{equation}
\small
    \mathcal{L}=\mathbb{E}_{z_0,\mI,\mhI,\tau,\mhD,\mbD,\mathbf{s},\epsilon}\big\|\epsilon-\epsilon_{\theta_\mathrm{v},\theta_\mathrm{c}}(\mI, \mhI,z_\tau,\tau,\mhD,\mbD,\mathbf{s})\big\|^2_2,
\end{equation}
where $\epsilon\sim \mathcal{N}(0,1)$ is the diffusion noise. 

\subsection{Dynamic Scene Update}
\label{sec:approach_step2}
Now that we have an outpainted video $\mI^{(1)}$ at the new pose $\bPi^{(1)}$, the next step is to produce a new dynamic point cloud and merge it with the previous one $\calP^{(0)}$. 

\myparagraph{Point cloud merging.} 
We exploit the depth model \cite{20pami/rene_midas} to obtain the video depth maps $\mD^{(1)}$ of the outpainted video $\mI^{(1)}$. Noting that there is a depth inconsistency problem between the estimated depths and the rendered ray depth map $\mhD^{(1)}$ in the mask regions $\mhM^{(1)}$, we follow common practices \cite{24cvpr/yu_wonderjourney,24arxiv/yu_wonderworld} to align the video depth maps $\mD^{(1)}$ with the ray depth maps $\mhD^{(1)}$ by finetuning the depth estimation model. After obtaining the aligned depth maps from the finetuned depth model, we follow \secref{sec:approach_step0} to initialize foreground point cloud $\calP^\mathrm{(1,f)}$ and background point cloud $\calP^\mathrm{(1,b)}$. Then we merge the new point clouds with the previous one to update the dynamic scenes,
\begin{equation}
    \calP^{(1)} = \calP^\mathrm{(1,f)} \cup \calP^\mathrm{(1,b)} \cup \calP^{(0)},
\end{equation}
where $\calP^{(1)}$ is the updated point cloud at the pose $\bPi^{(1)}$. In this way, we could generate perpetual dynamic scenes along fly-through camera trajectories by looping \secref{sec:approach_step1} and \secref{sec:approach_step2} with new camera poses and scene prompts.

\myparagraph{Background completion.} In practice, we find that the missing background point clouds occluded by the moving foreground in the scene cause inconsistency when rendering along fly-through camera trajectories (as shown in \figref{fig:ablation_4d_representation}). We hence propose to employ our video outpainting model to generate background videos of the occluded parts. After that, we extract depth maps of the background videos and reconstruct the background point clouds with the extracted depth maps. Details are in \supp. We also update the scene with these point clouds.

\CheckRmv{
\begin{figure}[t]
  \centering
   \includegraphics[width=\linewidth]{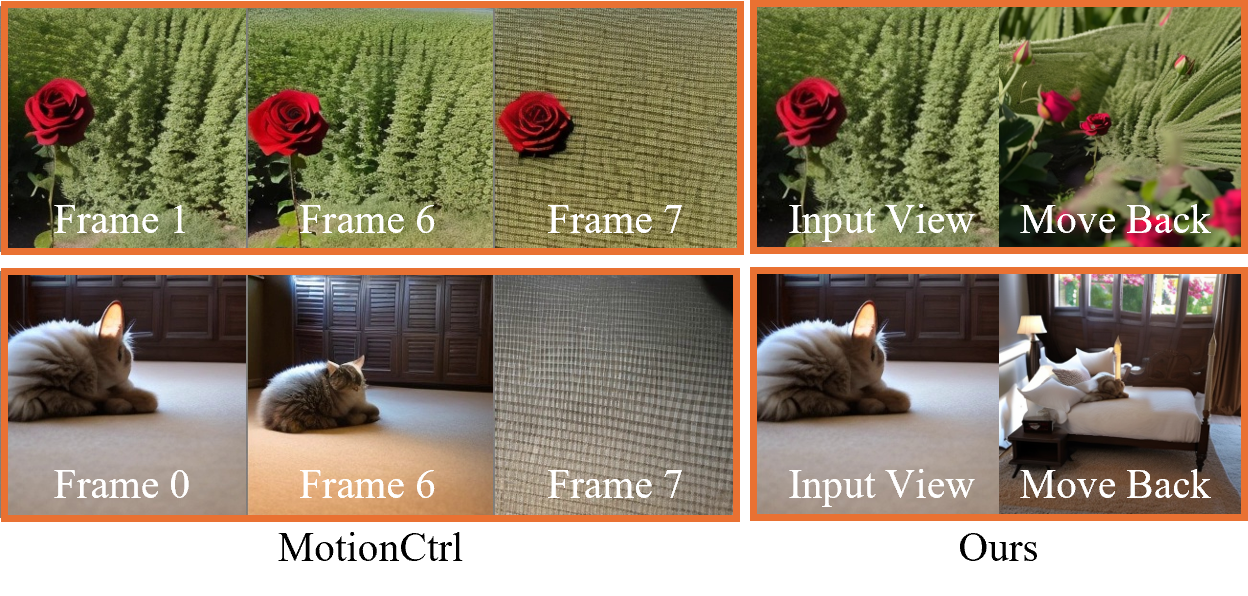}
\vspace{-7mm}
   \caption{Visual comparisons with video diffusion model MotionCtrl \cite{24siggraph/wang_motionctrl}. Full details are in \supp. \ourframework outpaints the scenes with large camera motions.}
   \label{fig:motionctrl_comp}
\end{figure}
}

\section{Experiments}
In this section, we first describe the details of our experimental setup, we then discuss the results of the scene generation with fly-through cameras and controllable scene generation. Finally, we conduct ablation studies of our model.

\subsection{Experiment Setups}
\paragraph{Data.} We trained our outpainting model with the OpenVid dataset \cite{25iclr/nan_openvid}, which contains high-resolution videos with detailed video prompts. In order to train with daily dynamic scene videos, we sampled 5,000 videos that contain outdoor dynamic scenes, including natural scenes (such as waterfalls, rivers and moving clouds) and urban scenes (such as walking people and moving cars). Additional data curation details can be found in the \supp. 

\begin{table}[t]
\caption{Quantitative results of controllable dynamic scene generation on the dynamic degree (DD), factual consistency (FC) \cite{24emnlp/he_videoscore} and CLIP \cite{21icml/alec_clip} scores. While WonderJourney \cite{24cvpr/yu_wonderjourney} only controls static scene generation with scene prompts, our model achieves controllable dynamic scene generation with higher performance.}
\label{tab:controllable_comp}
\resizebox{\columnwidth}{!}{%
\begin{tabular}{l|cc}
\toprule
methods           & WonderJourney \cite{24cvpr/yu_wonderjourney}              & ours                         \\\midrule
CLIP-SIM $\uparrow$ / DD $\uparrow$ / FC $\uparrow$ & 24.98 / 2.23 / 1.30 & \textbf{25.23} / \textbf{3.13} / \textbf{1.32} \\ \bottomrule
\end{tabular}%
}
\end{table}

\myparagraph{Training details.} We employed the LoRA \cite{22iclr/hu_lora} to finetune the \texttt{CogvideoX-5B-I2V} model \cite{24iclr/yang_cogvideox}. The LoRA rank was $256$ and the learning rate was $1\times 10^{-4}$ with a cosine schedule. The video resolution is 16$\times$512$\times$512, and hence $N=16$ and $h=w=512$. For calculating the ray information, we used the depth model \cite{20pami/rene_midas} to extract the depth map of each video frame. Then, we followed the \sArt video outpainting model MOTIA \cite{24eccv/wang_motia} and exploited the same mask proportion for outpainting.

\myparagraph{Camera details.}
To set up dynamic scenes with fly-through camera trajectories, we followed the common practices in \cite{24cvpr/yu_wonderjourney,253dv/popov_camctrl3d,24eccv/kong_dreamdrone}, where cameras move along a straight line or rotate. In the former case, we moved the camera backward by 0.0005 units between two adjacent camera poses, where one unit corresponds to the normalized 3D coordinate defined in PyTorch3D, and in the latter case, we rotated the camera by 0.45 radians. We generated camera paths by linearly interpolating translation and approximating the rotation using uniform angular steps.
\CheckRmv{
\begin{figure}[t]
  \centering
   \includegraphics[width=\linewidth]{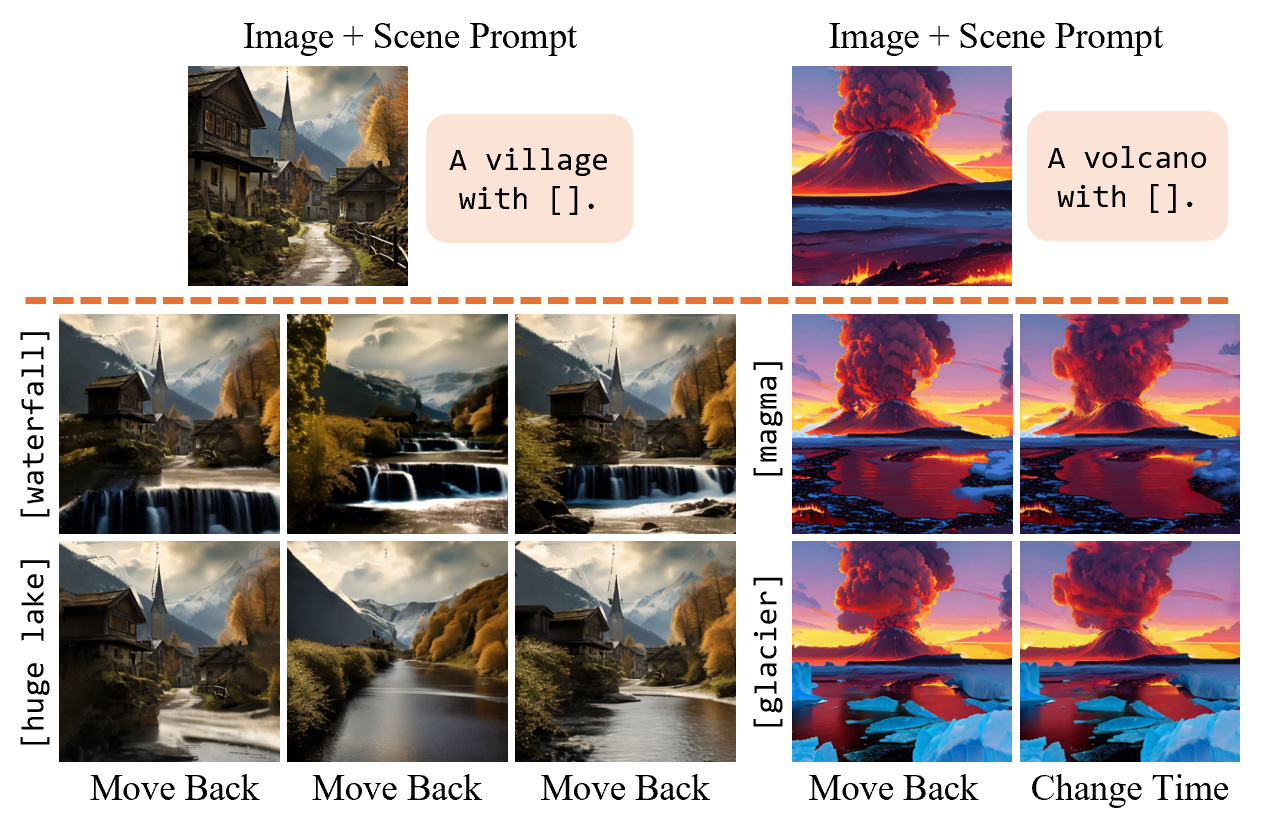}
   \caption{Controllable scene generation from input images with scene prompts. Full details are shown in \supp. \ourframework successfully controls the dynamic scene outpainting content with the corresponding scene prompt.}
   \label{fig:control_gen}
\end{figure}
}

\subsection{Scene Generation with Fly-through Cameras}
\figref{fig:large_cam_motion} shows the dynamic scene generation results with long fly-through trajectories. We tested our model on both the background dynamics (water flowing in the waterfall) and the foreground dynamics (person walking on the path). Compared to the initial point cloud, our model generated perpetual dynamic scenes by exploiting the proposed scene outpainting process. As there is no direct baseline for perpetual dynamic scene generation, we compare our method with the perpetual static scene generation method WonderJourney \cite{24cvpr/yu_wonderjourney} in \figref{fig:view_interpolation}, \sArt~4D scene generation model CAT4D \cite{24arxiv/cat4d} in \figref{fig:prev_failure} and camera-controllable video diffusion model MotionCtrl \cite{24siggraph/wang_motionctrl} in \figref{fig:motionctrl_comp}. These figures demonstrate that our model can better generate perpetual dynamic scenes with 3D consistent motions by exploiting our ray outpainting model.

\subsection{Scene Generation Controlled by Text}
To test the controllable dynamic scene generation ability of our model, we generated dynamic scenes from village and volcano images with scene prompts. As shown in \figref{fig:control_gen}, for the village image, we control the outpainting content by introducing two different text prompts: \texttt{waterfall} and \texttt{huge lake}. For the volcano image, we prompted the model to generate \texttt{glacier} and \texttt{magma}. Our model successfully controls the perpetual dynamic scene generation with the given scene prompts.  Following the common practices in \cite{24nips/yu_4real,24eccv/kong_dreamdrone,24cvpr/yu_wonderjourney,24arxiv/yu_wonderworld}, we employed the CLIP \cite{21icml/alec_clip} scores between the novel view images and the input scene prompts and the dynamic degree (DD) and factual consistency (FC) indices proposed by VideoScore \cite{24emnlp/he_videoscore} to thoroughly evaluate the generation quality of the video. \tabref{tab:controllable_comp} shows that, compared to WonderJourney, our model achieves better performance by controlling dynamic scene generation.

\begin{table}[t]
  \centering
\caption{Quantitative results of 3D consistent motions. VQ, TC and TA respectively denote the visual quality, temporal consistency and text-to-video alignment matrices \cite{24emnlp/he_videoscore}. Our model achieves better performance by achieving 3D consistent motions.}
\label{tab:3d_consistency}
\resizebox{\columnwidth}{!}{%
\begin{tabular}{l|ccccc}
\toprule
methods      & CLIP-SIM $\uparrow$ & VQ $\uparrow$    & TC $\uparrow$   & TA $\uparrow$   & FC $\uparrow$      \\ \midrule
MOTIA  \cite{24eccv/wang_motia}   & 23.80      & 2.281 & 1.852 & 2.656 & 1.859  \\
w/o ray info & 23.12  & 2.562 & 1.984 & 2.672 & 1.977 \\
\ourframework & \textbf{24.70} & \textbf{2.578} & \textbf{2.234} & \textbf{2.938} & \textbf{2.188}  \\ \bottomrule
\end{tabular}%
}
\end{table}

\begin{table}[t]
  \centering
\caption{Quantitative ablation studies of the proposed foreground layer (\S \ref{sec:approach_step0}) and background completion (\S \ref{sec:approach_step2}) strategy.}
\label{tab:ablation}

\resizebox{\columnwidth}{!}{%
\begin{tabular}{l|ccccc}
\toprule
methods           & CLIP-SIM $\uparrow$ & VQ $\uparrow$    & TC $\uparrow$    & TA $\uparrow$    & FC $\uparrow$    \\ \midrule
w/o fg layer      & 23.88    & 1.891 & 0.871 & 2.219 & 1.102  \\
w/o bg completion & 24.62    & 1.906 & 0.871 & 2.547 & 1.195 \\
\ourframework    & \textbf{25.48}    & \textbf{1.977} & \textbf{1.031} & \textbf{2.562} & \textbf{1.320}  \\\bottomrule
\end{tabular}%
}

\end{table}

\subsection{Ablation Study}
\myparagraph{Ray outpainting.} To fairly evaluate the proposed ray outpainting model, we conducted experiments on five scenes containing common dynamic objects: cars, people, water and clouds. For each scene, we sampled 16 novel view images from the generated dynamic point cloud at 16 different timestamps along the fly-through camera trajectories. As there are no 3D video outpainting baselines, we chose the 2D outpainting model MOTIA \cite{24eccv/wang_motia} for comparisons. \figref{fig:ablation_ray} shows that by introducing the ray information for outpainting, our model successfully outpaints 3D consistent motions at the outpainting boundaries. In this way, our ray outpainting model achieves better performance in \tabref{tab:3d_consistency}.

\myparagraph{4D representation.} We tested the proposed foreground representation and background completion strategy in \figref{fig:ablation_4d_representation} and \tabref{tab:ablation}. It can be seen that our model renders consistent foregrounds with foreground layers and generates a plausible background with the background completion strategy.

\section{Conclusion}
We proposed \ourframework, a new model for perpetual dynamic scene generation,  which reformulates this task as an outpainting problem. By considering pixels as rays and  conditioning on ray depth and distance maps, \ourframework can generate scenes with 3D consistent motions. Experimental results demonstrate the superiority of our model.

\myparagraph{Limitations.} Challenges remain in handling reflections, shadows, fine structures, multi-view rendering, and depth discontinuities, which can be explored in future work.

\section*{Acknowledgments} The authors sincerely thank Prof. Kostas Daniilidis, Dr. Jiahui Lei, Qiao Feng, Chen Wang, Ziqing Xu, and Uday Kiran for their generous and insightful feedback on this work. We also acknowledge the support of a Penn Engineering graduate fellowship, Penn startup funds, and the Research Collaboration on the Mathematical and Scientific Foundations of Deep Learning under grants NSF 2031985 and Simons 814201.

{
    \small
    \bibliographystyle{ieeenat_fullname}
    \bibliography{main}

\begin{thebibliography}{60}
\providecommand{\natexlab}[1]{#1}
\providecommand{\url}[1]{\texttt{#1}}
\expandafter\ifx\csname urlstyle\endcsname\relax
  \providecommand{\doi}[1]{doi: #1}\else
  \providecommand{\doi}{doi: \begingroup \urlstyle{rm}\Url}\fi

\bibitem[Bahmani et~al.(2025)Bahmani, Skorokhodov, Siarohin, Menapace, Qian, Vasilkovsky, Lee, Wang, Zou, Tagliasacchi, et~al.]{25iclr/sherwin_vd3d}
Sherwin Bahmani, Ivan Skorokhodov, Aliaksandr Siarohin, Willi Menapace, Guocheng Qian, Michael Vasilkovsky, Hsin-Ying Lee, Chaoyang Wang, Jiaxu Zou, Andrea Tagliasacchi, et~al.
\newblock {VD3D}: Taming large video diffusion transformers for {3D} camera control.
\newblock In \emph{ICLR}, 2025.

\bibitem[Bautista et~al.(2022)Bautista, Guo, Abnar, Talbott, Toshev, Chen, Dinh, Zhai, Goh, Ulbricht, et~al.]{22nips/miguel_gaudi}
Miguel~Angel Bautista, Pengsheng Guo, Samira Abnar, Walter Talbott, Alexander Toshev, Zhuoyuan Chen, Laurent Dinh, Shuangfei Zhai, Hanlin Goh, Daniel Ulbricht, et~al.
\newblock Gaudi: A neural architect for immersive {3D} scene generation.
\newblock In \emph{NeurIPS}, 2022.

\bibitem[Broxton et~al.(2020)Broxton, Flynn, Overbeck, Erickson, Hedman, Duvall, Dourgarian, Busch, Whalen, and Debevec]{20tog/michael_immersive}
Michael Broxton, John Flynn, Ryan Overbeck, Daniel Erickson, Peter Hedman, Matthew Duvall, Jason Dourgarian, Jay Busch, Matt Whalen, and Paul Debevec.
\newblock Immersive light field video with a layered mesh representation.
\newblock In \emph{ToG}, 2020.

\bibitem[Collet et~al.(2015)Collet, Chuang, Sweeney, Gillett, Evseev, Calabrese, Hoppe, Kirk, and Sullivan]{15tog/alvaro_freevideo}
Alvaro Collet, Ming Chuang, Pat Sweeney, Don Gillett, Dennis Evseev, David Calabrese, Hugues Hoppe, Adam Kirk, and Steve Sullivan.
\newblock High-quality streamable free-viewpoint video.
\newblock In \emph{ToG}, 2015.

\bibitem[Deitke et~al.(2023)Deitke, Schwenk, Salvador, Weihs, Michel, VanderBilt, Schmidt, Ehsani, Kembhavi, and Farhadi]{23cvpr/matt_objaverse}
Matt Deitke, Dustin Schwenk, Jordi Salvador, Luca Weihs, Oscar Michel, Eli VanderBilt, Ludwig Schmidt, Kiana Ehsani, Aniruddha Kembhavi, and Ali Farhadi.
\newblock Objaverse: A universe of annotated 3d objects.
\newblock In \emph{CVPR}, 2023.

\bibitem[DeVries et~al.(2021)DeVries, Bautista, Srivastava, Taylor, and Susskind]{21iccv/terrance_unconstrained}
Terrance DeVries, Miguel~Angel Bautista, Nitish Srivastava, Graham~W Taylor, and Joshua~M Susskind.
\newblock Unconstrained scene generation with locally conditioned radiance fields.
\newblock In \emph{ICCV}, 2021.

\bibitem[Gan et~al.(2021)Gan, Schwartz, Alter, Schrimpf, Traer, De~Freitas, Kubilius, Bhandwaldar, Haber, Sano, et~al.]{21nips/gan_tdw}
C Gan, J Schwartz, S Alter, M Schrimpf, J Traer, J De~Freitas, J Kubilius, A Bhandwaldar, N Haber, M Sano, et~al.
\newblock {ThreeDWorld}: A platform for interactive multi-modal physical simulation.
\newblock In \emph{NeurIPS}, 2021.

\bibitem[Gao* et~al.(2024)Gao*, Holynski*, Henzler, Brussee, Martin-Brualla, Srinivasan, Barron, and Poole*]{24nips/gao_cat3d}
Ruiqi Gao*, Aleksander Holynski*, Philipp Henzler, Arthur Brussee, Ricardo Martin-Brualla, Pratul~P. Srinivasan, Jonathan~T. Barron, and Ben Poole*.
\newblock {CAT3D}: Create anything in {3D} with multi-view diffusion models.
\newblock In \emph{NeurIPS}, 2024.

\bibitem[Gu et~al.(2023)Gu, Xiang, Li, Ling, Liu, Mu, Tang, Tao, Wei, Yao, Yuan, Xie, Huang, Chen, and Su]{23iclr/gu_maniskill}
Jiayuan Gu, Fanbo Xiang, Xuanlin Li, Zhan Ling, Xiqiang Liu, Tongzhou Mu, Yihe Tang, Stone Tao, Xinyue Wei, Yunchao Yao, Xiaodi Yuan, Pengwei Xie, Zhiao Huang, Rui Chen, and Hao Su.
\newblock {ManiSkill2}: A unified benchmark for generalizable manipulation skills.
\newblock In \emph{ICLR}, 2023.

\bibitem[Guo et~al.(2024)Guo, Yang, Rao, Liang, Wang, Qiao, Agrawala, Lin, and Dai]{24iclr/guo_animatediff}
Yuwei Guo, Ceyuan Yang, Anyi Rao, Zhengyang Liang, Yaohui Wang, Yu Qiao, Maneesh Agrawala, Dahua Lin, and Bo Dai.
\newblock {AnimateDiff}: Animate your personalized text-to-image diffusion models without specific tuning.
\newblock In \emph{ICLR}, 2024.

\bibitem[He et~al.(2025)He, Xu, Guo, Wetzstein, Dai, Li, and Yang]{25iclr/he_camctrl}
Hao He, Yinghao Xu, Yuwei Guo, Gordon Wetzstein, Bo Dai, Hongsheng Li, and Ceyuan Yang.
\newblock {CameraCtrl}: Enabling camera control for text-to-video generation.
\newblock In \emph{ICLR}, 2025.

\bibitem[He et~al.(2024)He, Jiang, Zhang, Ku, Soni, Siu, Chen, Chandra, Jiang, Arulraj, Wang, Do, Ni, Lyu, Narsupalli, Fan, Lyu, Lin, and Chen]{24emnlp/he_videoscore}
Xuan He, Dongfu Jiang, Ge Zhang, Max Ku, Achint Soni, Sherman Siu, Haonan Chen, Abhranil Chandra, Ziyan Jiang, Aaran Arulraj, Kai Wang, Quy~Duc Do, Yuansheng Ni, Bohan Lyu, Yaswanth Narsupalli, Rongqi Fan, Zhiheng Lyu, Yuchen Lin, and Wenhu Chen.
\newblock {VideoScore}: Building automatic metrics to simulate fine-grained human feedback for video generation.
\newblock In \emph{EMNLP}, 2024.

\bibitem[H{\"o}llein et~al.(2023)H{\"o}llein, Cao, Owens, Johnson, and Nie{\ss}ner]{23iccv/lukas_t2room}
Lukas H{\"o}llein, Ang Cao, Andrew Owens, Justin Johnson, and Matthias Nie{\ss}ner.
\newblock {Text2Room}: Extracting textured {3D} meshes from {2D} text-to-image models.
\newblock In \emph{ICCV}, 2023.

\bibitem[Hong et~al.(2022)Hong, Ding, Zheng, Liu, and Tang]{22iclr/hong_cogvideo}
Wenyi Hong, Ming Ding, Wendi Zheng, Xinghan Liu, and Jie Tang.
\newblock {CogVideo}: Large-scale pretraining for text-to-video generation via transformers.
\newblock In \emph{ICLR}, 2022.

\bibitem[Hou et~al.(2025)Hou, Wei, Zeng, and Chen]{25iclr/hou_camctrlfree}
Chen Hou, Guoqiang Wei, Yan Zeng, and Zhibo Chen.
\newblock Training-free camera control for video generation.
\newblock In \emph{ICLR}, 2025.

\bibitem[Hu et~al.(2022)Hu, Shen, Wallis, Allen-Zhu, Li, Wang, Wang, Chen, et~al.]{22iclr/hu_lora}
Edward~J Hu, Yelong Shen, Phillip Wallis, Zeyuan Allen-Zhu, Yuanzhi Li, Shean Wang, Lu Wang, Weizhu Chen, et~al.
\newblock {LoRA}: Low-rank adaptation of large language models.
\newblock In \emph{ICLR}, 2022.

\bibitem[Hu et~al.(2021)Hu, Ravi, Berg, and Pathak]{21iccv/hu_worldsheet}
Ronghang Hu, Nikhila Ravi, Alexander~C Berg, and Deepak Pathak.
\newblock Worldsheet: Wrapping the world in a 3d sheet for view synthesis from a single image.
\newblock In \emph{ICCV}, 2021.

\bibitem[Jain et~al.(2023)Jain, Li, Chiu, Hassani, Orlov, and Shi]{23cvpr/jitesh_oneformer}
Jitesh Jain, Jiachen Li, Mang~Tik Chiu, Ali Hassani, Nikita Orlov, and Humphrey Shi.
\newblock {OneFormer}: One transformer to rule universal image segmentation.
\newblock In \emph{CVPR}, 2023.

\bibitem[Kerbl et~al.(2023)Kerbl, Kopanas, Leimk{\"u}hler, and Drettakis]{23tog/bernhard_3dgs}
Bernhard Kerbl, Georgios Kopanas, Thomas Leimk{\"u}hler, and George Drettakis.
\newblock {3D} gaussian splatting for real-time radiance field rendering.
\newblock In \emph{TOG}, 2023.

\bibitem[Kirillov et~al.(2023)Kirillov, Mintun, Ravi, Mao, Rolland, Gustafson, Xiao, Whitehead, Berg, Lo, et~al.]{23iccv/alexander_sam}
Alexander Kirillov, Eric Mintun, Nikhila Ravi, Hanzi Mao, Chloe Rolland, Laura Gustafson, Tete Xiao, Spencer Whitehead, Alexander~C Berg, Wan-Yen Lo, et~al.
\newblock Segment anything.
\newblock In \emph{ICCV}, 2023.

\bibitem[Kong et~al.(2024)Kong, Lian, Mi, and Wang]{24eccv/kong_dreamdrone}
Hanyang Kong, Dongze Lian, Michael~Bi Mi, and Xinchao Wang.
\newblock {DreamDrone}: Text-to-image diffusion models are zero-shot perpetual view generators.
\newblock In \emph{ECCV}, 2024.

\bibitem[Kuang et~al.(2024)Kuang, Cai, He, Xu, Li, Guibas, and Wetzstein]{24nips/kuang_collcamctrl}
Zhengfei Kuang, Shengqu Cai, Hao He, Yinghao Xu, Hongsheng Li, Leonidas Guibas, and Gordon Wetzstein.
\newblock Collaborative video diffusion: Consistent multi-video generation with camera control.
\newblock In \emph{NeurIPS}, 2024.

\bibitem[Labs(2024)]{flux2024}
Black~Forest Labs.
\newblock Flux.
\newblock \url{https://github.com/black-forest-labs/flux}, 2024.

\bibitem[Labs et~al.(2025)Labs, Batifol, Blattmann, Boesel, Consul, Diagne, Dockhorn, English, English, Esser, Kulal, Lacey, Levi, Li, Lorenz, Müller, Podell, Rombach, Saini, Sauer, and Smith]{25arxiv/flux}
Black~Forest Labs, Stephen Batifol, Andreas Blattmann, Frederic Boesel, Saksham Consul, Cyril Diagne, Tim Dockhorn, Jack English, Zion English, Patrick Esser, Sumith Kulal, Kyle Lacey, Yam Levi, Cheng Li, Dominik Lorenz, Jonas Müller, Dustin Podell, Robin Rombach, Harry Saini, Axel Sauer, and Luke Smith.
\newblock Flux.1 kontext: Flow matching for in-context image generation and editing in latent space, 2025.

\bibitem[Lei et~al.(2023)Lei, Tang, and Jia]{23cvpr/lei_rgbd2}
Jiabao Lei, Jiapeng Tang, and Kui Jia.
\newblock {RGBD2}: Generative scene synthesis via incremental view inpainting using {RGBD} diffusion models.
\newblock In \emph{CVPR}, 2023.

\bibitem[Li et~al.(2025)Li, Pan, Yang, Xu, Zhou, Zhang, Li, Kadambi, Wang, Tu, et~al.]{25iclr/4k4dgen}
Renjie Li, Panwang Pan, Bangbang Yang, Dejia Xu, Shijie Zhou, Xuanyang Zhang, Zeming Li, Achuta Kadambi, Zhangyang Wang, Zhengzhong Tu, et~al.
\newblock {4K4DGen}: Panoramic {4D} generation at {4K} resolution.
\newblock In \emph{ICLR}, 2025.

\bibitem[Li et~al.(2022)Li, Wang, Snavely, and Kanazawa]{22eccv/li_infinitzero}
Zhengqi Li, Qianqian Wang, Noah Snavely, and Angjoo Kanazawa.
\newblock {Infinitenature-Zero}: Learning perpetual view generation of natural scenes from single images.
\newblock In \emph{ECCV}, 2022.

\bibitem[Ling et~al.(2024{\natexlab{a}})Ling, Kim, Torralba, Fidler, and Kreis]{24cvpr/ling_ayg}
Huan Ling, Seung~Wook Kim, Antonio Torralba, Sanja Fidler, and Karsten Kreis.
\newblock Align your gaussians: Text-to-{4D} with dynamic {3D} gaussians and composed diffusion models.
\newblock In \emph{CVPR}, 2024{\natexlab{a}}.

\bibitem[Ling et~al.(2024{\natexlab{b}})Ling, Sheng, Tu, Zhao, Xin, Wan, Yu, Guo, Yu, Lu, et~al.]{24cvpr/lu_dl3dv}
Lu Ling, Yichen Sheng, Zhi Tu, Wentian Zhao, Cheng Xin, Kun Wan, Lantao Yu, Qianyu Guo, Zixun Yu, Yawen Lu, et~al.
\newblock {DL3DV-10K}: A large-scale scene dataset for deep learning-based {3D} vision.
\newblock In \emph{CVPR}, 2024{\natexlab{b}}.

\bibitem[Liu et~al.(2021)Liu, Tucker, Jampani, Makadia, Snavely, and Kanazawa]{21iccv/liu_infinite}
Andrew Liu, Richard Tucker, Varun Jampani, Ameesh Makadia, Noah Snavely, and Angjoo Kanazawa.
\newblock {Infinite Nature}: Perpetual view generation of natural scenes from a single image.
\newblock In \emph{ICCV}, 2021.

\bibitem[Liu et~al.(2024)Liu, Lin, Zeng, Long, Liu, Komura, and Wang]{24iclr/liu_syncdreamer}
Yuan Liu, Cheng Lin, Zijiao Zeng, Xiaoxiao Long, Lingjie Liu, Taku Komura, and Wenping Wang.
\newblock {SyncDreamer}: Generating multiview-consistent images from a single-view image.
\newblock In \emph{ICLR}, 2024.

\bibitem[Mildenhall et~al.(2020)Mildenhall, Srinivasan, Tancik, Barron, Ramamoorthi, and Ng]{20eccv/Ben_nerf}
Ben Mildenhall, Pratul~P. Srinivasan, Matthew Tancik, Jonathan~T. Barron, Ravi Ramamoorthi, and Ren Ng.
\newblock {{NeRF}}: Representing scenes as neural radiance fields for view synthesis.
\newblock In \emph{ECCV}, 2020.

\bibitem[Nan et~al.(2025)Nan, Xie, Zhou, Fan, Yang, Chen, Li, Yang, and Tai]{25iclr/nan_openvid}
Kepan Nan, Rui Xie, Penghao Zhou, Tiehan Fan, Zhenheng Yang, Zhijie Chen, Xiang Li, Jian Yang, and Ying Tai.
\newblock {Openvid-1M}: A large-scale high-quality dataset for text-to-video generation.
\newblock In \emph{ICLR}, 2025.

\bibitem[Pan et~al.(2024)Pan, Yang, Zhu, and Zhang]{24arxiv/pan_efficient4d}
Zijie Pan, Zeyu Yang, Xiatian Zhu, and Li Zhang.
\newblock {Efficient4D}: Fast dynamic {3D} object generation from a single-view video.
\newblock \emph{arXiv preprint arXiv 2401.08742}, 2024.

\bibitem[Popov et~al.(2025)Popov, Raj, Krainin, Li, Freeman, and Rubinstein]{253dv/popov_camctrl3d}
Stefan Popov, Amit Raj, Michael Krainin, Yuanzhen Li, William~T. Freeman, and Michael Rubinstein.
\newblock {CamCtrl3D}: Single-image scene exploration with precise {3D} camera control.
\newblock In \emph{3DV}, 2025.

\bibitem[Radford et~al.(2021)Radford, Kim, Hallacy, Ramesh, Goh, Agarwal, Sastry, Askell, Mishkin, Clark, et~al.]{21icml/alec_clip}
Alec Radford, Jong~Wook Kim, Chris Hallacy, Aditya Ramesh, Gabriel Goh, Sandhini Agarwal, Girish Sastry, Amanda Askell, Pamela Mishkin, Jack Clark, et~al.
\newblock Learning transferable visual models from natural language supervision.
\newblock In \emph{ICML}, 2021.

\bibitem[Ranftl et~al.(2020)Ranftl, Lasinger, Hafner, Schindler, and Koltun]{20pami/rene_midas}
Ren{\'e} Ranftl, Katrin Lasinger, David Hafner, Konrad Schindler, and Vladlen Koltun.
\newblock Towards robust monocular depth estimation: Mixing datasets for zero-shot cross-dataset transfer.
\newblock In \emph{IEEE TPAMI}, 2020.

\bibitem[Ren et~al.(2023)Ren, Pan, Tang, Zhang, Cao, Zeng, and Liu]{23arxiv/ren_dreamgs4d}
Jiawei Ren, Liang Pan, Jiaxiang Tang, Chi Zhang, Ang Cao, Gang Zeng, and Ziwei Liu.
\newblock {{DreamGaussian4D}}: Generative {4D} gaussian splatting.
\newblock \emph{arXiv preprint arXiv:2312.17142}, 2023.

\bibitem[Ren et~al.(2025)Ren, Shen, Huang, Ling, Lu, Nimier-David, Müller, Keller, Fidler, and Gao]{25cvpr/ren_gen3c}
Xuanchi Ren, Tianchang Shen, Jiahui Huang, Huan Ling, Yifan Lu, Merlin Nimier-David, Thomas Müller, Alexander Keller, Sanja Fidler, and Jun Gao.
\newblock {GEN3C}: 3d-informed world-consistent video generation with precise camera control.
\newblock In \emph{CVPR}, 2025.

\bibitem[Rombach et~al.(2022{\natexlab{a}})Rombach, Blattmann, Lorenz, Esser, and Ommer]{22cvpr/robin_sd}
Robin Rombach, Andreas Blattmann, Dominik Lorenz, Patrick Esser, and Bj{\"o}rn Ommer.
\newblock High-resolution image synthesis with latent diffusion models.
\newblock In \emph{CVPR}, 2022{\natexlab{a}}.

\bibitem[Rombach et~al.(2022{\natexlab{b}})Rombach, Blattmann, Lorenz, Esser, and Ommer]{22cvpr/rombach_ldm}
Robin Rombach, Andreas Blattmann, Dominik Lorenz, Patrick Esser, and Bj{\"o}rn Ommer.
\newblock High-resolution image synthesis with latent diffusion models.
\newblock In \emph{CVPR}, pages 10684--10695, 2022{\natexlab{b}}.

\bibitem[Sun et~al.(2024)Sun, Chen, Liu, Chen, Duan, Zhang, and Wang]{24arxiv/sun_dimensionx}
Wenqiang Sun, Shuo Chen, Fangfu Liu, Zilong Chen, Yueqi Duan, Jun Zhang, and Yikai Wang.
\newblock {DimensionX}: Create any {3D and 4D} scenes from a single image with controllable video diffusion.
\newblock \emph{arXiv preprint arXiv:2411.04928}, 2024.

\bibitem[Tian et~al.(2023)Tian, Du, and Duan]{23iccv/tian_mononerf}
Fengrui Tian, Shaoyi Du, and Yueqi Duan.
\newblock {MonoNeRF}: Learning a generalizable dynamic radiance field from monocular videos.
\newblock In \emph{ICCV}, 2023.

\bibitem[Tian et~al.(2024)Tian, Duan, Wang, Guo, and Du]{24iclr/tian_semflow}
Fengrui Tian, Yueqi Duan, Angtian Wang, Jianfei Guo, and Shaoyi Du.
\newblock {Semantic Flow}: Learning semantic fields of dynamic scenes from monocular videos.
\newblock In \emph{ICLR}, 2024.

\bibitem[Van~Hoorick et~al.(2024)Van~Hoorick, Wu, Ozguroglu, Sargent, Liu, Tokmakov, Dave, Zheng, and Vondrick]{24eccv/van_gencamdolly}
Basile Van~Hoorick, Rundi Wu, Ege Ozguroglu, Kyle Sargent, Ruoshi Liu, Pavel Tokmakov, Achal Dave, Changxi Zheng, and Carl Vondrick.
\newblock Generative camera dolly: Extreme monocular dynamic novel view synthesis.
\newblock In \emph{ECCV}, 2024.

\bibitem[Wang et~al.(2024{\natexlab{a}})Wang, Wu, Huang, Shi, Shen, Song, Liu, and Li]{24eccv/wang_motia}
Fu-Yun Wang, Xiaoshi Wu, Zhaoyang Huang, Xiaoyu Shi, Dazhong Shen, Guanglu Song, Yu Liu, and Hongsheng Li.
\newblock Be-your-outpainter: Mastering video outpainting through input-specific adaptation.
\newblock In \emph{ECCV}, 2024{\natexlab{a}}.

\bibitem[Wang et~al.(2024{\natexlab{b}})Wang, Leroy, Cabon, Chidlovskii, and Revaud]{24cvpr/wang_dust3r}
Shuzhe Wang, Vincent Leroy, Yohann Cabon, Boris Chidlovskii, and Jerome Revaud.
\newblock {DUSt3R}: Geometric {3D} vision made easy.
\newblock In \emph{CVPR}, 2024{\natexlab{b}}.

\bibitem[Wang et~al.(2023)Wang, Yuan, Wang, Li, Chen, Xia, Luo, and Shan]{24siggraph/wang_motionctrl}
Zhouxia Wang, Ziyang Yuan, Xintao Wang, Yaowei Li, Tianshui Chen, Menghan Xia, Ping Luo, and Ying Shan.
\newblock Motionctrl: A unified and flexible motion controller for video generation.
\newblock In \emph{ACM SIGGRAPH}, 2023.

\bibitem[Wu et~al.(2024)Wu, Gao, Poole, Trevithick, Zheng, Barron, and Holynski]{24arxiv/cat4d}
Rundi Wu, Ruiqi Gao, Ben Poole, Alex Trevithick, Changxi Zheng, Jonathan~T Barron, and Aleksander Holynski.
\newblock {Cat4D}: Create anything in {4D} with multi-view video diffusion models.
\newblock \emph{arXiv preprint arXiv:2411.18613}, 2024.

\bibitem[Xie et~al.(2024)Xie, Yao, Voleti, Jiang, and Jampani]{24arxiv/xie_sv4d}
Yiming Xie, Chun-Han Yao, Vikram Voleti, Huaizu Jiang, and Varun Jampani.
\newblock {SV4D}: Dynamic {3D} content generation with multi-frame and multi-view consistency.
\newblock \emph{arXiv preprint arXiv:2407.17470}, 2024.

\bibitem[Xing et~al.(2024)Xing, Xia, Zhang, Chen, Yu, Liu, Liu, Wang, Shan, and Wong]{24eccv/xing_dyncrafter}
Jinbo Xing, Menghan Xia, Yong Zhang, Haoxin Chen, Wangbo Yu, Hanyuan Liu, Gongye Liu, Xintao Wang, Ying Shan, and Tien-Tsin Wong.
\newblock {Dynamicrafter}: Animating open-domain images with video diffusion priors.
\newblock In \emph{ICLR}, 2024.

\bibitem[Yang et~al.(2024{\natexlab{a}})Yang, Yang, Gupta, Han, Fei-Fei, and Xie]{24arxiv/yang_thinkspc}
Jihan Yang, Shusheng Yang, Anjali~W. Gupta, Rilyn Han, Li Fei-Fei, and Saining Xie.
\newblock {Thinking in Space}: How multimodal large language models see, remember and recall spaces.
\newblock \emph{arXiv preprint arXiv:2412.14171}, 2024{\natexlab{a}}.

\bibitem[Yang et~al.(2024{\natexlab{b}})Yang, Hou, Huang, Ma, Wan, Zhang, Chen, and Liao]{24siggraph/yang_directavideo}
Shiyuan Yang, Liang Hou, Haibin Huang, Chongyang Ma, Pengfei Wan, Di Zhang, Xiaodong Chen, and Jing Liao.
\newblock {Direct-a-Video}: Customized video generation with user-directed camera movement and object motion.
\newblock In \emph{SIGGRAPH}, 2024{\natexlab{b}}.

\bibitem[Yang et~al.(2024{\natexlab{c}})Yang, Teng, Zheng, Ding, Huang, Xu, Yang, Hong, Zhang, Feng, et~al.]{24iclr/yang_cogvideox}
Zhuoyi Yang, Jiayan Teng, Wendi Zheng, Ming Ding, Shiyu Huang, Jiazheng Xu, Yuanming Yang, Wenyi Hong, Xiaohan Zhang, Guanyu Feng, et~al.
\newblock {CogVideoX}: Text-to-video diffusion models with an expert transformer.
\newblock In \emph{ICLR}, 2024{\natexlab{c}}.

\bibitem[Yin et~al.(2023)Yin, Xu, Wang, Zhao, and Wei]{23arxiv/yin_4dgen}
Yuyang Yin, Dejia Xu, Zhangyang Wang, Yao Zhao, and Yunchao Wei.
\newblock {{4DGen}}: Grounded {4D} content generation with spatial-temporal consistency.
\newblock \emph{arXiv preprint arXiv:2312.17225}, 2023.

\bibitem[Yu et~al.(2024{\natexlab{a}})Yu, Wang, Zhuang, Menapace, Siarohin, Cao, Jeni, Tulyakov, and Lee]{24nips/yu_4real}
Heng Yu, Chaoyang Wang, Peiye Zhuang, Willi Menapace, Aliaksandr Siarohin, Junli Cao, Laszlo~A Jeni, Sergey Tulyakov, and Hsin-Ying Lee.
\newblock {4Real}: Towards photorealistic {4D} scene generation via video diffusion models.
\newblock In \emph{NeurIPS}, 2024{\natexlab{a}}.

\bibitem[Yu et~al.(2024{\natexlab{b}})Yu, Duan, Herrmann, Freeman, and Wu]{24arxiv/yu_wonderworld}
Hong-Xing Yu, Haoyi Duan, Charles Herrmann, William~T Freeman, and Jiajun Wu.
\newblock {WonderWorld}: Interactive {3D} scene generation from a single image.
\newblock \emph{arXiv preprint arXiv:2406.09394}, 2024{\natexlab{b}}.

\bibitem[Yu et~al.(2024{\natexlab{c}})Yu, Duan, Hur, Sargent, Rubinstein, Freeman, Cole, Sun, Snavely, Wu, et~al.]{24cvpr/yu_wonderjourney}
Hong-Xing Yu, Haoyi Duan, Junhwa Hur, Kyle Sargent, Michael Rubinstein, William~T Freeman, Forrester Cole, Deqing Sun, Noah Snavely, Jiajun Wu, et~al.
\newblock {WonderJourney}: Going from anywhere to everywhere.
\newblock In \emph{CVPR}, 2024{\natexlab{c}}.

\bibitem[Zhang et~al.(2023)Zhang, Rao, and Agrawala]{23iccv/zhang_controlnet}
Lvmin Zhang, Anyi Rao, and Maneesh Agrawala.
\newblock Adding conditional control to text-to-image diffusion models.
\newblock In \emph{ICCV}, 2023.

\bibitem[Zhao et~al.(2025)Zhao, Lin, Lin, Yan, Li, Yang, Wang, Lee, and Wang]{25iclr/zhao_genxd}
Yuyang Zhao, Chung-Ching Lin, Kevin Lin, Zhiwen Yan, Linjie Li, Zhengyuan Yang, Jianfeng Wang, Gim~Hee Lee, and Lijuan Wang.
\newblock {GenXD}: Generating any {3D} and {4D} scenes.
\newblock In \emph{ICLR}, 2025.

\end{thebibliography}
}
\newpage
\clearpage

\noindent We organize the supplementary material as follows. 
\begin{itemize}[wide,itemindent=5pt]
    \item \secref{sec:video} provides an additional video for better visualization of the generated dynamic scenes, implementation code of \ourframework, and a table of notations used in the paper. 
    \item \secref{sec:impl_details} presents more implementation details of the proposed ray outpainting model and 4D point cloud reconstruction. 
    \item \secref{sec:bg_completion} introduces the details of the background completion strategy.
    \item \secref{sec:more_results}  presents more scene generation results and details of camera-controllable video diffusion model MotionCtrl \cite{24siggraph/wang_motionctrl}.
\end{itemize}

\begin{table*}[t]
\centering
\caption{Notations.}
\label{tab:notations}
\begin{tabular}{cl}
\toprule
\textbf{Notation} & \textbf{Description} \\
\toprule
$\bPi^{(i)}$ & Camera pose for the $i^\mathrm{th}$ view \\
$\mI^{(i)}$ & Video frames at pose $\bPi^{(i)}$: $\{\mI^{(i)}_t\in \sR^{h\times w \times 3}\}_{t=0}^{N-1}$ \\
$\mI^{(i)}_t$ & Video frame at timestamp $t$ and pose $\bPi^{(i)}$ \\
$\mD^{(i)}$ & Depth maps at pose $\bPi^{(i)}$: $\{\mD^{(i)}_t\in \mathbb{R}^{h\times w}\}_{t=0}^{N-1}$ \\
$\mD^{(i)}_t$ & Depth map at timestamp $t$ and pose $\bPi^{(i)}$ \\
$\mM^{(i)}$ & Binary foreground masks at pose $\bPi^{(i)}$: $\{\mM^{(i)}_t\in \{0,1\}^{h\times w}\}_{t=0}^{N-1}$ \\
$\mM^{(i)}_t$ & Binary foreground mask at timestamp $t$ and pose $\bPi^{(i)}$ \\
$\mathcal{P}^{(i)}$ & 4D point cloud constructed until pose $\bPi^{(i)}$ \\
$\mathcal{P}^{(i)}_t$ & 4D point cloud at timestamp $t$ constructed until pose $\bPi^{(i)}$ \\
$\vp $ & Point in 4D point cloud: $\vp= (\vx, t, \vc)$ with position $\vx$, timestamp $t$, color $\vc$ \\
$\vx$ & 3D position of a point, $\vx \in \mathbb{R}^3$ \\
$\vc$ & Color of a point, $\vc \in \mathbb{R}^3$ \\
$\calP^{(i,\mathrm{f})},\calP^{(i,\mathrm{b})}$ & Foreground, background point cloud until pose $\bPi^{(i)}$ \\
$\mI^{(i,\mathrm{f})},\mI^{(i,\mathrm{b})}$ & Foreground, background video frames at pose $\bPi^{(i)}$ \\
$\mD^{(i,\mathrm{f})},\mD^{(i,\mathrm{b})}$ & Foreground, background depth maps at pose $\bPi^{(i)}$ \\
$\phi$ & Mapping function from 2D image and depth to 3D point cloud \\
$\varphi$ & Image rasterization function \\
$\mhI^{(i)}$ & Rasterized (partial) video at pose $\bPi^{(i)}$ given point cloud $\mathcal{P}^{(i-1)}$\\
$\mhD^{(i)}$ & Rasterized ray depth maps at pose $\bPi^{(i)}$ given point cloud $\mathcal{P}^{(i-1)}$\\
$\mhM^{(i)}$ & Binary masks for observed regions in rasterization at pose $\bPi^{(i)}$\\
$\vr^{(i)}(x,y)$ & Unit-norm ray vector from camera origin at pose $\bPi^{(i)}$ to pixel $(x,y)$ \\
$\vo^{(i)}$ & Camera center of pose $\bPi^{(i)}$ \\
$\operatorname{dist}_{\mathrm{r2p}}$ & Distance function between a ray and a point \\
$\mbD^{(i)}_t$ & Distance map between rays and point cloud at timestamp $t$ \\
$\mathbf{s}$ & Scene prompt describing desired motion \\
$z_\tau$ & Noisy video at noise step $\tau$ \\
$\theta_\mathrm{v}$ & Training parameters of video diffusion model \\
$\theta_\mathrm{c}$ & Training parameters of ControlNet \\
\bottomrule
\end{tabular}
\end{table*}
\section{Video, Code, and Notation Table}
\label{sec:video}
We encourage readers to watch the video in the supplementary material to better understand our visualization results.
Moreover, we provide the implementation code in the supplementary material; our code and pretrained models will be publicly available. For convenience, we summarize the notations in the paper in \tabref{tab:notations}.
\section{Implementation Details}
\label{sec:impl_details}
We present further implementation details and the dataset filtering process here.

\myparagraph{Scene generation details.}
Inspired by WonderJourney \cite{24cvpr/yu_wonderjourney}, we adjust the depth maps with SAM \cite{23iccv/alexander_sam} to promote spatio-temporal consistency of the depth for each object in the scene.

\myparagraph{Dataset filtering process.}
We used the keywords in \tabref{tab:filter_keywords} to filter videos in OpenVid \cite{25iclr/nan_openvid}. After filtering, we randomly selected $5\times 10^3$ videos to fine-tune our video outpainting models.
\begin{table*}[ht]
\caption{Filter keywords.}
\label{tab:filter_keywords}
\resizebox{\linewidth}{!}{%
\begin{tabular}{l|c}
\toprule
 &
  mountain, forest, river, lake, ocean,    canyon,  meadow, hill, city, trees, cloud, wind,   nature \\ &grassland, beach, snow, waterfall, stream, wildlife, pond, dunes, island, rainforest, sunset, sunrise, storm,   rain, thunderstorm\\ keywords &snowstorm, lightning, rainbow, twilight, dawn, dusk,   star, aurora,
  moon, downtown, skyline, street, road, pedestrians, crowd\\  & transport, bridge, tower, skyscraper, urban, bustling, bus, metropolis, 
  coast, beachfront, harbor, port, pier, boat, yacht\\ &
  dock, marina, waterfront,   seaside, swimming, surf, sailboat,
  sun, shore, tidal, waves, bay, lagoon,   village, town, countryside\\ &rural,  greenhouse, sunflower field,   landscape, sky,  lake, water, sea, seas, cityscape, pedestrian,   pedestrians, cars, traffic, people 
  \\\bottomrule
\end{tabular}%
}
\end{table*}

\section{Background Completion}
\label{sec:bg_completion}
As described in the method section of the main paper, in practice, we find that background regions occluded by the dynamic foregrounds can become exposed when rendering the dynamic scene with a fly-through camera trajectory. Since these regions are fully occluded by the foregrounds under camera pose $\mathbf{\Pi}^\mathrm{(i)}$, the missing parts cause inconsistent colors in 3D when rendering the dynamic scene with fly-through cameras. To address this issue, we propose to employ our outpainting model to fill the missing colors in the background. More specifically, after obtaining the fixed viewpoint video $\mI^\mathrm{(i)}$, the depth maps $\mD^\mathrm{(i)}$ and foreground masks $\mM^\mathrm{(i)}$, we obtain the background video $\mI^\mathrm{(i,b)}$ by employing foreground masks on the video $\mI^\mathrm{(i)}$. For the regions occluded by the foreground in the video $\mI^\mathrm{(i,b)}$, we sample the ray information following (5), (6) and (7) in the main paper. Then we exploit our ray outpainting model to inpaint the video colors at the occluded regions with the sampled ray information. After that, we obtain the background depths of these regions with a depth estimation model \cite{20pami/rene_midas} and reconstruct the background point cloud on these regions. Finally, we merge these occluded background point clouds into our dynamic scene point cloud.

\CheckRmv{
\begin{figure*}[ht]
  \centering
   \includegraphics[width=\linewidth]{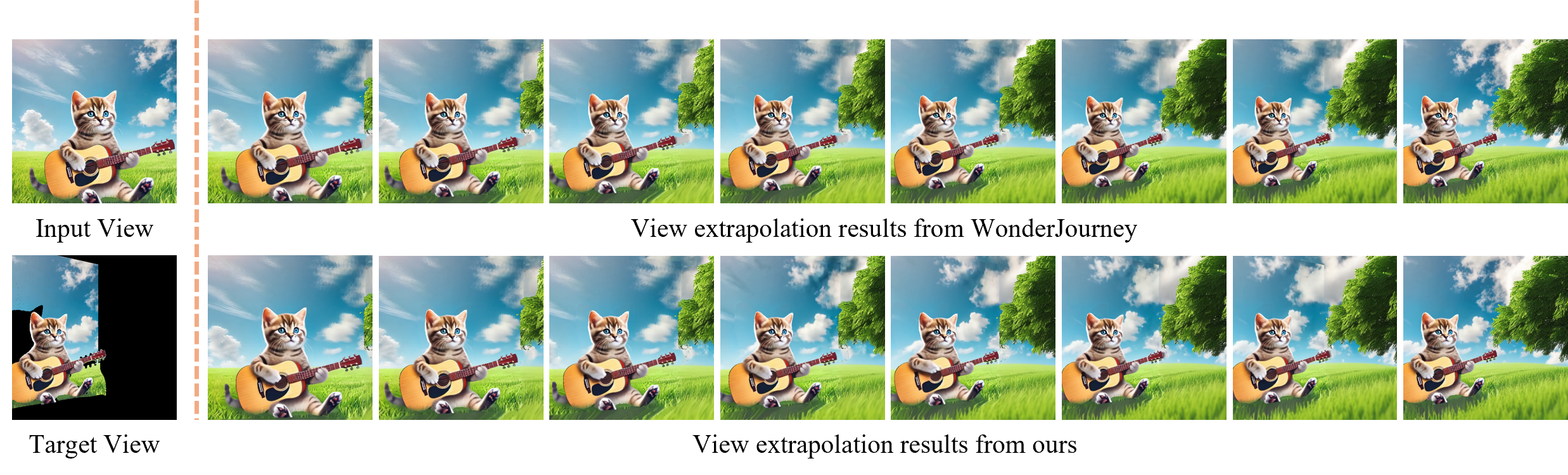}

   \caption{Space-time interpolation comparisons between \ourframework and WonderJourney \cite{24cvpr/yu_wonderjourney}.}
   \label{fig:space_time_interpolate_supp}
\end{figure*}
}

\CheckRmv{
\begin{figure*}[ht]
  \centering
   \includegraphics[width=\linewidth]{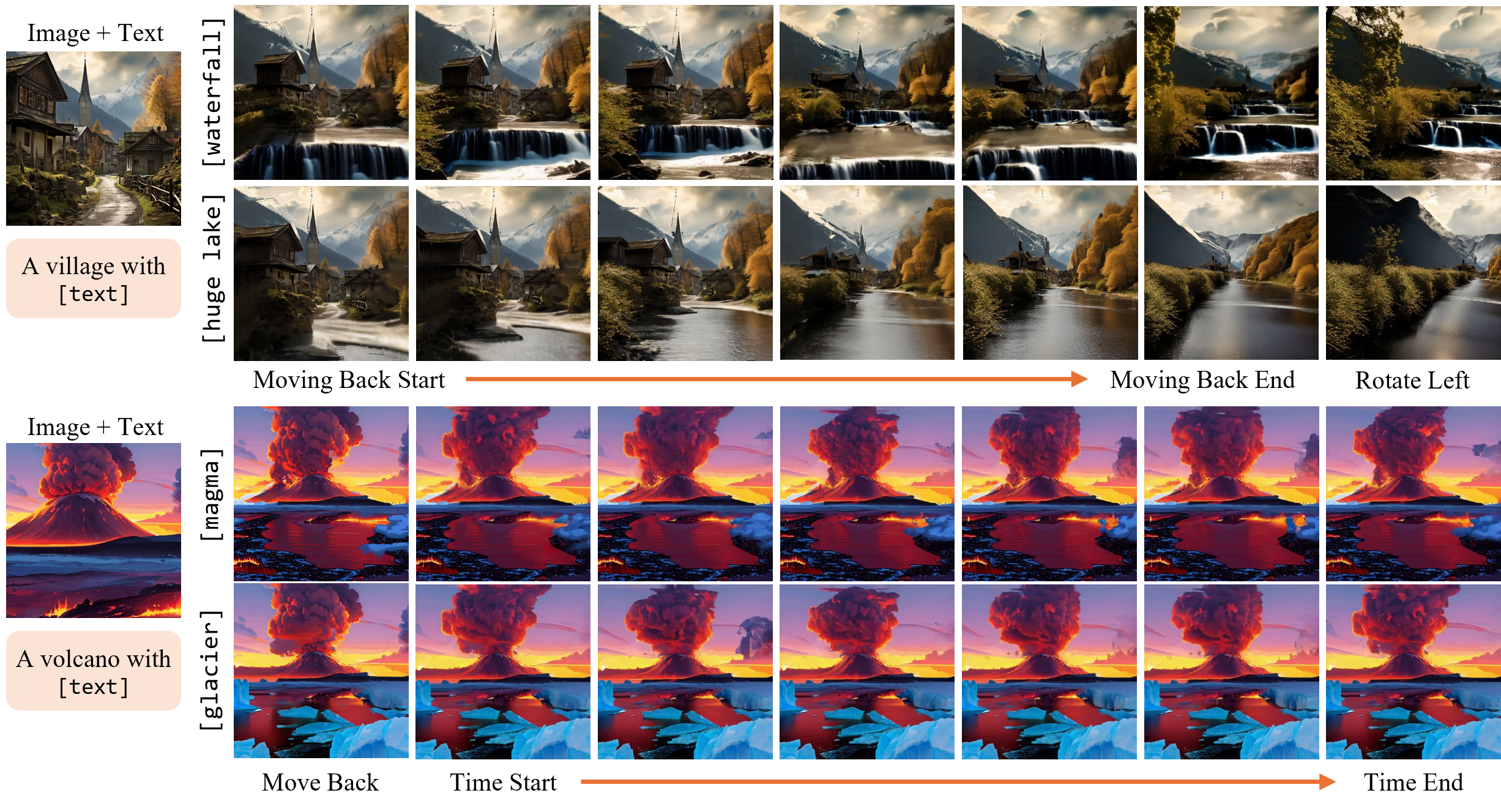}

   \caption{Controllable scene generation from input images with scene prompts. Given an image as input, \ourframework successfully controls the dynamic scene outpainting content with the corresponding text prompt.}
   \label{fig:control_gen_full_supp}
\end{figure*}
}

\CheckRmv{
\begin{figure*}[!t]
  \centering
   \includegraphics[width=\linewidth]{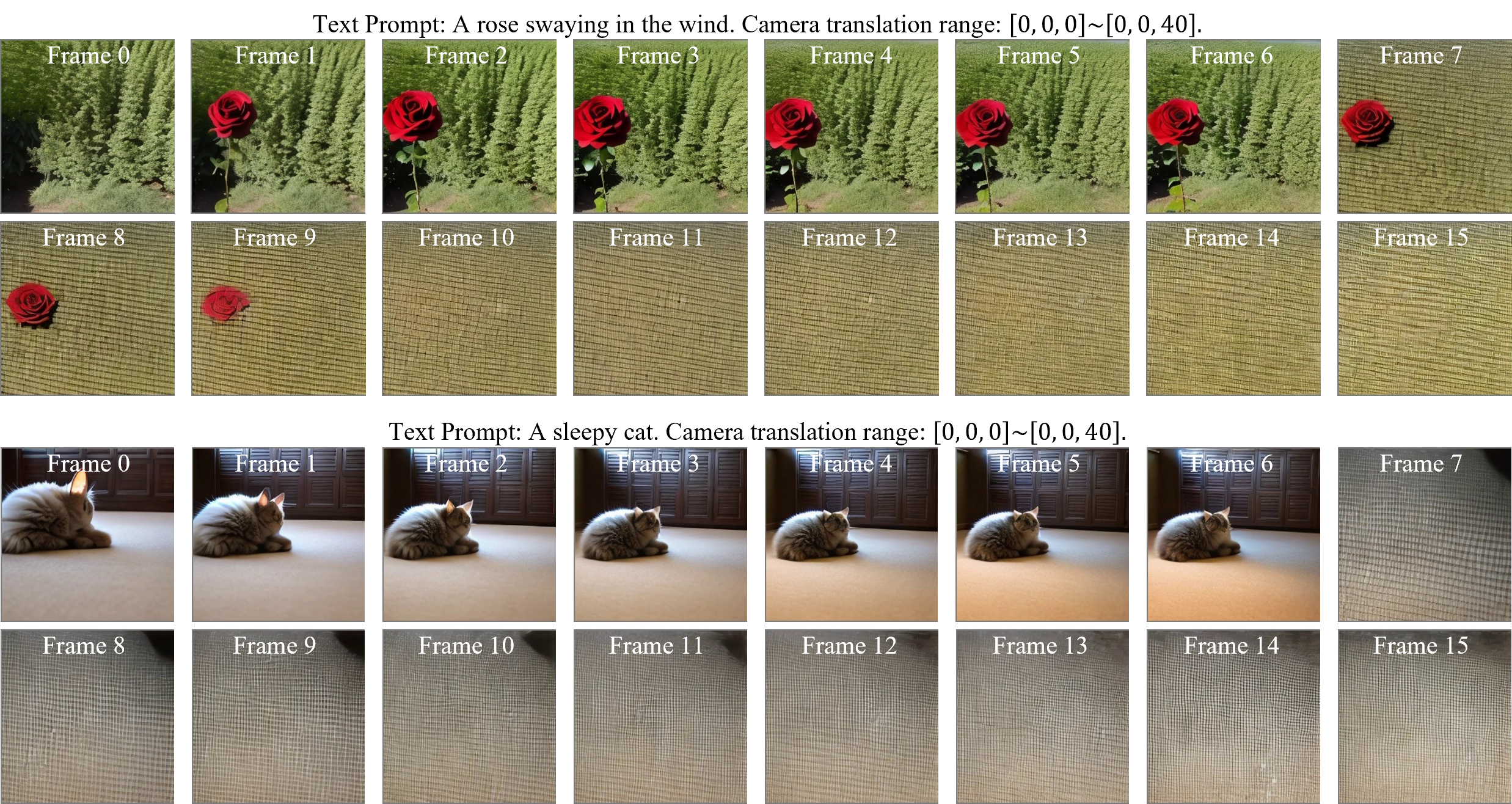}

   \caption{Detailed results of MotionCtrl \cite{24siggraph/wang_motionctrl} with large camera translation inputs. It can be seen that the model fails to generate videos with large camera motions.}
   \label{fig:motionctrl}
\end{figure*}
}

\section{More Dynamic Scene Generation Results}
\label{sec:more_results}
\myparagraph{Space-time interpolation.} \figref{fig:space_time_interpolate_supp} presents visualizations of space-time interpolation results. In this figure, a cartoon cat is playing guitar with moving clouds. It can be seen that our model successfully renders dynamic scenes with fly-through cameras, while WonderJourney \cite{24cvpr/yu_wonderjourney} only interpolates in the static scene.

\myparagraph{Controllable scene generation.}  \figref{fig:control_gen_full_supp} reports additional results on controllable scene generation. We control the village scene to generate \texttt{cascade waterfall} and \texttt{huge lake} separately. It can be seen that our model successfully controls scene generations with fly-through cameras.

\myparagraph{MotionCtrl comparison details.} We chose to compare with MotionCtrl \cite{24siggraph/wang_motionctrl}, which is a camera-controllable video diffusion model presented in the figure of the main paper. As MotionCtrl only generates videos from text prompts, we use the first frame generated by MotionCtrl as the input of our model. To test the camera control performance in the MotionCtrl model, we evenly sample 16 camera translations from $[0,0,0]$ to $[0,0,40]$ and introduce them into MotionCtrl models. The results are shown in \figref{fig:motionctrl}. As MotionCtrl learns from training videos with relatively limited camera movements, it fails to generate videos with large camera motions.

\end{document}